\documentclass[10pt,twocolumn,letterpaper]{article}

\usepackage[pagenumbers]{cvpr} % To force page numbers, e.g. for an arXiv version

\usepackage{graphicx}
\usepackage{amsmath}
\usepackage{amssymb}
\usepackage{booktabs}

\usepackage[pagebackref,breaklinks,colorlinks]{hyperref}

\usepackage{multirow}
\usepackage{subcaption}
\usepackage{wrapfig}
\usepackage{enumitem}
\usepackage{etoolbox}
\usepackage{subfiles}

\newcommand{\MyMapTemplatePrefixc}[4]{\expandafter#1\csname#3#4\endcsname{#2{#4}}} % it remembles a template: \#3#4 --> #2{#4}
\forcsvlist{\MyMapTemplatePrefixc {\def} {\mathcal}{c}} {A,B,C,D,E,F,G,H,I,J,K,L,M,N,O,P,Q,R,S,T,U,V,W,X,Y,Z}  % e.g., \cA --> \mathcal{A}
\newcommand{\MyMapTemplateNoPrefix}[3]{\expandafter#1\csname#3\endcsname{#2{#3}}}
\forcsvlist{\MyMapTemplateNoPrefix {\def} {\mathbf} } {0,1,a,b,c,d,e, f, g, h, i, j, k, l, m, n, o, p, q, r, u, v, w, x, y, z} % e.g., \a --> \mathbf{a}
\forcsvlist{\MyMapTemplateNoPrefix {\def} {\mathbf} } {A,B,C,D,E,F,G,H,I,J,K,L,M,N,O,P,Q,R,S,T,U,V,W,X,Y,Z}  % e.g., \A --> \mathcal{A}

\usepackage[capitalize]{cleveref}
\crefname{section}{Sec.}{Secs.}
\Crefname{section}{Section}{Sections}
\Crefname{table}{Table}{Tables}
\crefname{table}{Tab.}{Tabs.}

\begin{document}

%%%%%%%%% TITLE - PLEASE UPDATE
\title{One-Shot Domain Adaptive and Generalizable Semantic Segmentation with Class-Aware Cross-Domain Transformers}

% \author{Rui Gong\\
% Institution1\\
% Institution1 address\\
% {\tt\small firstauthor@i1.org}
% % For a paper whose authors are all at the same institution,
% % omit the following lines up until the closing ``}''.
% % Additional authors and addresses can be added with ``\and'',
% % just like the second author.
% % To save space, use either the email address or home page, not both
% % \and
% % Second Author\\
% % Institution2\\
% % First line of institution2 address\\
% % {\tt\small secondauthor@i2.org}
% \and
% Qin Wang
% \and
% Dengxin Dai
% \and
% Luc Van Gool
% }
\author{Rui Gong \textsuperscript{\rm 1}, Qin Wang \textsuperscript{\rm 1}, Dengxin Dai \textsuperscript{\rm 2}, Luc Van Gool \textsuperscript{\rm 1,3}\\
\textsuperscript{\rm 1} Computer Vision Lab, ETH Zurich, \textsuperscript{\rm 2} MPI for Informatics, \textsuperscript{\rm 3} VISICS, KU Leuven\\
{\tt\small\{gongr, qin.wang, vangool\}@vision.ee.ethz.ch, ddai@mpi-inf.mpg.de}
}
\maketitle

%%%%%%%%% ABSTRACT
\begin{abstract}
   Unsupervised sim-to-real domain adaptation (UDA) for semantic segmentation aims to improve the real-world test performance of a model trained on simulated data. It can save the cost of manually labeling data in real-world applications such as robot vision and autonomous driving. Traditional UDA often assumes that there are abundant unlabeled real-world data samples available during training for the adaptation. However, such an assumption does not always hold in practice owing to the collection difficulty and the scarcity of the data. Thus, we aim to relieve this need on a large number of real data, and explore the one-shot unsupervised sim-to-real domain adaptation (OSUDA) and generalization (OSDG) problem, where only one real-world data sample is available. To remedy the limited real data knowledge, we first construct the pseudo-target domain by stylizing the simulated data with the one-shot real data. To mitigate the sim-to-real domain gap on both the style and spatial structure level and facilitate the sim-to-real adaptation, we further propose to use class-aware cross-domain transformers with an intermediate domain randomization strategy to extract the domain-invariant knowledge, from both the simulated and pseudo-target data. We demonstrate the effectiveness of our approach for OSUDA and OSDG on different benchmarks, outperforming the state-of-the-art methods by a large margin, 10.87, 9.59, 13.05 and 15.91 mIoU on GTA, SYNTHIA$\rightarrow$Cityscapes, Foggy Cityscapes, respectively.
\end{abstract}

%%%%%%%%% BODY TEXT
\section{Introduction}\label{sec:introduction}

\begin{figure}
    \centering
    \includegraphics[width=\linewidth]{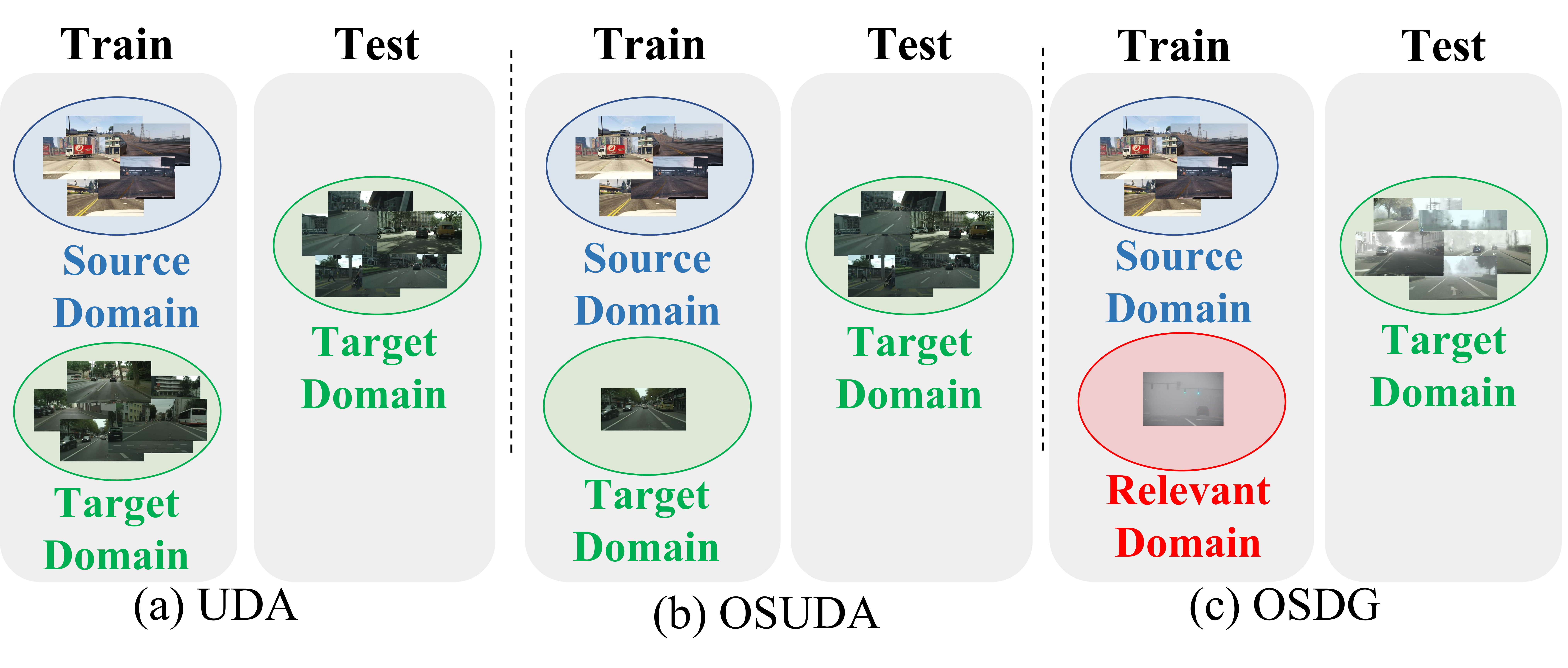}
    \vspace{-20pt}
    \caption{\textbf{Comparisons between UDA, OSUDA and OSDG.} UDA assumes there are a large number of unlabeled target real images available. Nevertheless, OSUDA and our \emph{proposed} OSDG only access one-shot target (target-relevant) domain image. }
    \vspace{-15pt}
    \label{fig:problem_setting}
\end{figure}

Semantic segmentation, which aims at assigning a semantic class label to each pixel of a given image, serves as an important and fundamental task in the field of computer vision. Deep leaning based semantic segmentation models~\cite{long2015fully,chen2017deeplab,cheng2021mask2former} have achieved great success recently, and are playing an important role in a slew of applications, \eg, robot vision and autonomous driving. Training these models typically relies on collecting and labeling a large amount of diverse real-world data, which is notoriously costly. In order to circumvent the problem, leveraging the power of simulation to automatically synthesize abundant labeled images serves as a promising way~\cite{richter2016playing,ros2016synthia,dosovitskiy2017carla}. However, the models trained on the simulated images often do not perform well on the real images, because of the distribution difference between the simulated and real images. To this end, the unsupervised sim-to-real domain adaptation (UDA) methods~\cite{Tsai_adaptseg_2018,tranheden2021dacs,zou2018unsupervised,hoffman2018cycada,vu2019advent} are developed recently to mitigate the domain gap between the well-labeled source domain, \emph{simulated images}, and the unlabeled target domain, \emph{real images}. Although much progress has been made, traditional UDA methods typically assume the access to a large number of real images, constituting the target domain. However, such an assumption does not always hold in real-world applications, since the collection of real-world images is not only costly but can also be very difficult due to the data scarcity, \eg, rare adverse weather and diseases. Thus, more recent works~\cite{choi2021robustnet,kim2021wedge,pan2018IBN-Net,yue2019domain,peng2022semantic,tobin2017domain,prakash2019structured} start to explore the more challenging problems of domain generalization (DG) and the domain randomization (DR), where the target domain images are assumed unavailable during training. While some progress is made, the DG/DR for semantic segmentation performance still falls far behind the UDA performance, suffering from the inaccessible target domain knowledge and limiting the sim-to-real applications.

Considering both the difficulty in data collection and the benefits from the guidance of the target domain, in this paper, we investigate the one-shot scenario, \ie, there is only one image available as the target domain, named as one-shot unsupervised sim-to-real domain adaptation (OSUDA). The setup aims to bridge the domain gap with only one-shot target guidance, which is generally available in real-world applications. Existing methods on OSUDA~\cite{luo2020adversarial,wu2021style} typically fall into two main categories, 1) by exploiting auxiliary style images, \eg, a separate set of images with diverse styles from ImageNet~\cite{deng2009imagenet}, to stylize the source domain images and search for the harder stylized samples around the one-shot target sample to improve the generality of the trained model, 2) by generating mixed-style images/ feature, \ie, intermediate domain, between the source and target domain, and utilize the consistency regularization between them to improve the robustness of the model when adapting to the target domain.  
Moreover, the domain gap between the source and target domain arises from the \emph{style} and \emph{spatial structure} differences~\cite{devaranjan2020meta,kar2019meta}. However, both the type 1) and 2) methods focus on reducing the domain gap on the style difference, but ignore the \emph{spatial structure} difference (\eg, the spatial location of the bus in Fig.~\ref{fig:crossformer}\textcolor{red}{a}).

We propose to improve the generalization ability on the target domain by taking both style and spatial structure differences into account. As assumed by~\cite{luo2020adversarial,wu2021style}, we first assume that the one-shot target sample contains style information which does not change drastically within the target dataset. 
To make full use of the one-shot target sample, we propose to firstly construct a \emph{pseudo-target domain} by stylizing the source domain images with the one-shot target domain image. This step bridges the style difference between the two domains. As the one-shot target sample does not necessarily provide the full picture of the spatial structure information in the entire target dataset, we propose to focus on the spatial structure information provided by the source domain. More specifically, we randomize the spatial structures from the source domain by class-mixed sampling strategy and synthesize a new domain called \emph{intermediate domain} . The intermediate domain contains randomized and augmented spatial structure information which can better generalized to the traget domain. We then design the \emph{class-aware cross-domain transformer} structure to facilitate the global alignment between the pseudo-target domain, source domain and intermediate domain. Compared to previous methods, our approach brings a few appealing benefits, (i) it does not require access to the large number of auxiliary data in type 1) methods, but still significantly outperforms the type 1) methods; (ii) it randomizes the intermediate domain to better mitigate the domain gap to the target domain on both the \emph{style} and \emph{spatial structure} level; (iii) it extracts the global domain-invariant knowledge among the source, intermediate and pseudo-target domain through the class-aware cross-domain transformer structure, leading to effective generalization to the target domain. Moreover, we extend the OSUDA task and relieve its constraint, by allowing the one-shot target image to be from a target-relevant domain instead of exactly the target domain, named as one-shot sim-to-real domain generalization (OSDG) problem. For example, in the GTA$\rightarrow$ Foggy Cityscapes setting, the proposed OSDG setup allows the one-shot target image, \ie, a foggy image, to be collected by randomly downloading from the internet, instead of exactly from Foggy Cityscapes dataset, further reducing the data collection difficulty.

Remarkably, extensive experiments on different benchmarks demonstrate that our method significantly outperforms the previous state-of-the-art (SOTA) methods on OSUDA and OSDG by a large margin, 10.87, 9.59, 13.05 and 15.91 mIoU on GTA, SYNTHIA$\rightarrow$ Cityscapes, Foggy Cityscapes, respectively.

\section{Related Work}
\textbf{Sim-to-Real Domain Adaptation/ Randomization/ Generalization.} \emph{Previous UDA works}~\cite{hoffman2018cycada,Tsai_adaptseg_2018,tranheden2021dacs,wulfmeier2017addressing,palazzo2020domain,keser2021content,gogoll2020unsupervised,zhang2019vr,bousmalis2018using,fang2018multi,wulfmeier2018incremental,messikommer2022bridging,yun2021target,sakuma2021geometry} assume there are abundant unlabeled real images in the target domain, which might not be the case in practice due to the collection difficulty, \eg, the rare adverse weather and diseases images. Instead, OSUDA and our proposed OSDG only require one-shot unlabeled image on the target/ target-relevant domain, which is more flexible and practical. From the method aspect, different from the \emph{semantic segmentation transformer based UDA method}~\cite{hoyer2022daformer} that only relies on the self-attention, our method proposes the class-aware cross-domain attention to further extract the domain-invariant knowledge. Besides, \emph{previous OSUDA/DR/DG methods}~\cite{luo2020adversarial,yue2019domain,kim2021wedge} typically need the access to auxiliary real images from other datasets, \eg, ImageNet~\cite{deng2009imagenet} and WikiArt~\cite{huang2017arbitrary}, increasing the GPU and storage memory requirement. However, our method only need one-shot target/relevant image, without relying on any other auxiliary data.

\textbf{Vision Transformers.} Transformer is originally proposed to model the sequence-to-sequence data in the filed of Natural Language Processing (NLP)~\cite{vaswani2017attention}. Recently, transformer models and their variants have demonstrated their effectiveness for different computer vision tasks, \eg, image classification~\cite{dosovitskiy2020vit,liu2021swin}, object detection~\cite{carion2020end,zhu2021deformable}, and semantic segmentation~\cite{zheng2021rethinking,strudel2021segmenter}. Among those works, the most related is~\cite{xu2022cdtrans} in terms of general methodology. The similarity is that, both \cite{xu2022cdtrans} and our method aim at improving the sim-to-real domain adaptation, with the help of cross-domain transformers. However, we have significant differences in the following aspects, 1) \cite{xu2022cdtrans} is designed for image classification, while we focus on semantic segmentation, yielding totally different framework structure and methods design; 2) The cross-domain attention in \cite{xu2022cdtrans} attends all image regions, while our class-aware cross-domain attention only attends regions of interest, \ie, non-sampled class regions.

\section{Method}
In OSUDA, we are given the well-labeled source domain dataset $\cD_s = \{\x_i^s, \y_i^s\}_{i=1}^{N_s}$, where $\x_i^s\in \mathbb{R}^{H\times W\times 3}$ is the simulated RGB color image, and $\y_i^s\in \mathbb{R}^{H\times W}$ is the corresponding semantic label map. $N$ is the number of semantic classes. In the target domain, we are given the one-shot real image, $\cD_t = \{\x^t\}$. $\cD_s$ and $\cD_t$ are from different distributions, \ie, $\x^s\sim P_s, \x^t\sim P_t$. The domain gap between $\cD_s$ and $\cD_t$, $P_s\neq P_t$, originates from the \emph{style} and \emph{spatial structure} difference between the simulated and real image. In OSDG, the one-shot target domain image is replaced with the one-shot target-relevant domain image, $\cD_r = \{\x^r\}$ (see Fig.~\ref{fig:problem_setting}), which is easy to be obtained. For example, the target-relevant domain image for Foggy Cityscapes dataset (target domain) can be easily obtained by, the user downloads one foggy image from the common search engine.

\begin{figure*}[t]
    \centering
    \includegraphics[width=\linewidth]{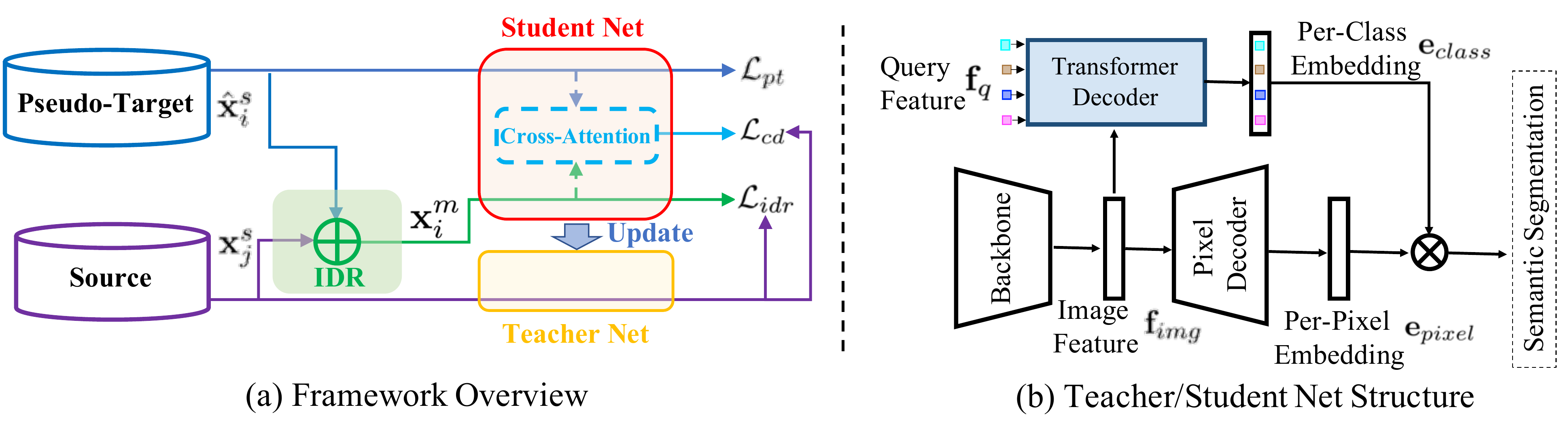}
    \vspace{-20pt}
    \caption{\textbf{Framework Overview and Network Structure.} (a) The whole mean-teacher framework adopts the pseudo-label based self-training strategy, where the pseudo-target domain construction, class mixed sampling based IDR, and global cross attention modules are developed. (b) shows the semantic segmentation network architecture, \ie, the teacher model $\cF_{\theta^{\prime}}$ and the student model $\cF_{\theta}$, which is composed of backbone, pixel decoder and transformer decoder.}
    \label{fig:framework}
    \vspace{-10pt}
\end{figure*}

\textbf{Method Overview.} The main difference between the one-shot setup and the general UDA setup is the further limited information on target domain, therefore it is critical to make full use of the limited information. While a single unlabeled target sample does not provide a lot of information about the data distribution of the entire target dataset, we assume that its style information is particularly useful and does not change drastically within the target dataset. To mitigate the domain gap on the \emph{style} level, we firstly make use of the style information from the one-shot target sample, and  construct the pseudo-target domain $\hat{\cD}_s = \{\hat{\x}_i^s\}_{i=1}^{N_s}$ (cf. Sec.~\ref{sec:pseudo_target}). Then, to compensate on the limited spatial structure information on the target dataset, we propose to randomize the source spatial structure to improve the generalization performance on spatial structure. To this end, we develop the class mixed sampling based intermediate domain randomization (IDR) training strategy to further reduce the domain gap on both \emph{style} level and \emph{spatial structure} level (cf. Sec.~\ref{sec:intermediate_dr}), where $\hat{\x}_i^s$, $\x_j^s$ and corresponding semantic label $\y_i^s$ and pseudo-label $\tilde{\y}_j^s$ are randomly mixed, respectively. Moreover, we propose the class-aware cross-domain transformers to extract the domain-invariant knowledge to facilitate the adaptation to the target domain (cf. Sec.~\ref{sec:mask_aware_trans}), by introducing the class-aware global cross-attention into the intermediate domain randomization training. The whole framework is shown in Fig.~\ref{fig:framework}.

\label{sec:train_strategy}\textbf{Training Strategy.} Following~\cite{tranheden2021dacs,olsson2021classmix}, we adopt the pseudo-label based self-training strategy, where the mean-teacher framework~\cite{tarvainen2017mean} (cf. Fig.~\ref{fig:framework}) is introduced. Both the teacher and the student model in the mean-teacher framework are semantic segmentation networks with the same structure. The teacher model, $\cF_{\theta^\prime}$, is used to output the final semantic prediction map. The student model, $\cF_{\theta}$, is used to backpropagate gradients and update weights $\theta$ based on all the relevant training loss. $\theta^\prime$ is an exponential moving average of $\theta$ throughout the optimization. Besides, the teacher model is also used to generate the pseudo-label $\tilde{\y}_j^s$ by feeding the source domain image sample $\x_j^s$, \ie, $\tilde{\y}_j^s = \cF_{\theta^\prime}(\x_j^s)$. During the inference stage, the final semantic segmentation map is obtained by $\cF_{\theta^\prime}(\x^t_{ts})$, where $\x^t_{ts}$ is the testing image.

\subsection{Pseudo-Target Domain for Style Alignment}\label{sec:pseudo_target}
In order to bridge the style difference between the source and the target domain, instead of directly aligning between the source and target domains, we propose to first construct a pseudo-target (PT) domain by augmenting the source with target style. For each source domain image ${\x}_i^s$, a pseudo-target sample $\hat{\x}_i^s$ will be generated according to the style of the one-shot target sample  $\hat{\x}_i^s = \cS(\x_i^s|\x^t)$, with the guidance of the one-shot target image $\x^t$, wehre $\cS(\cdot|\x^t)$ represents the image translation mapping conditioned on the style of $\x^t$. In order to prevent the overfitting to the one-shot target image and preserve the content, for $\cS(\cdot|\x^t)$, we adopt an off-the-shelf image translation framework~\cite{huang2018multimodal}  with the weighted perceptual loss~\cite{johnson2016perceptual} to generate the pseudo-target domain. By introducing the pseudo-target domain, we alleviate the style discrepancy between the domains and focus on the structural difference in other components. 
 With the generated pseudo-target domain, we employ the standard cross-entropy loss to learn the model, \ie, $\cL_{pt} = \sum_i CE(\cF_\theta(\hat{\x}_i^s), \y_i^s)$.

\subsection{Class Mixed Intermediate Domain Randomization for Spatial Structural Generalization}\label{sec:intermediate_dr}
\begin{figure*}
    \centering
    \includegraphics[width=\linewidth]{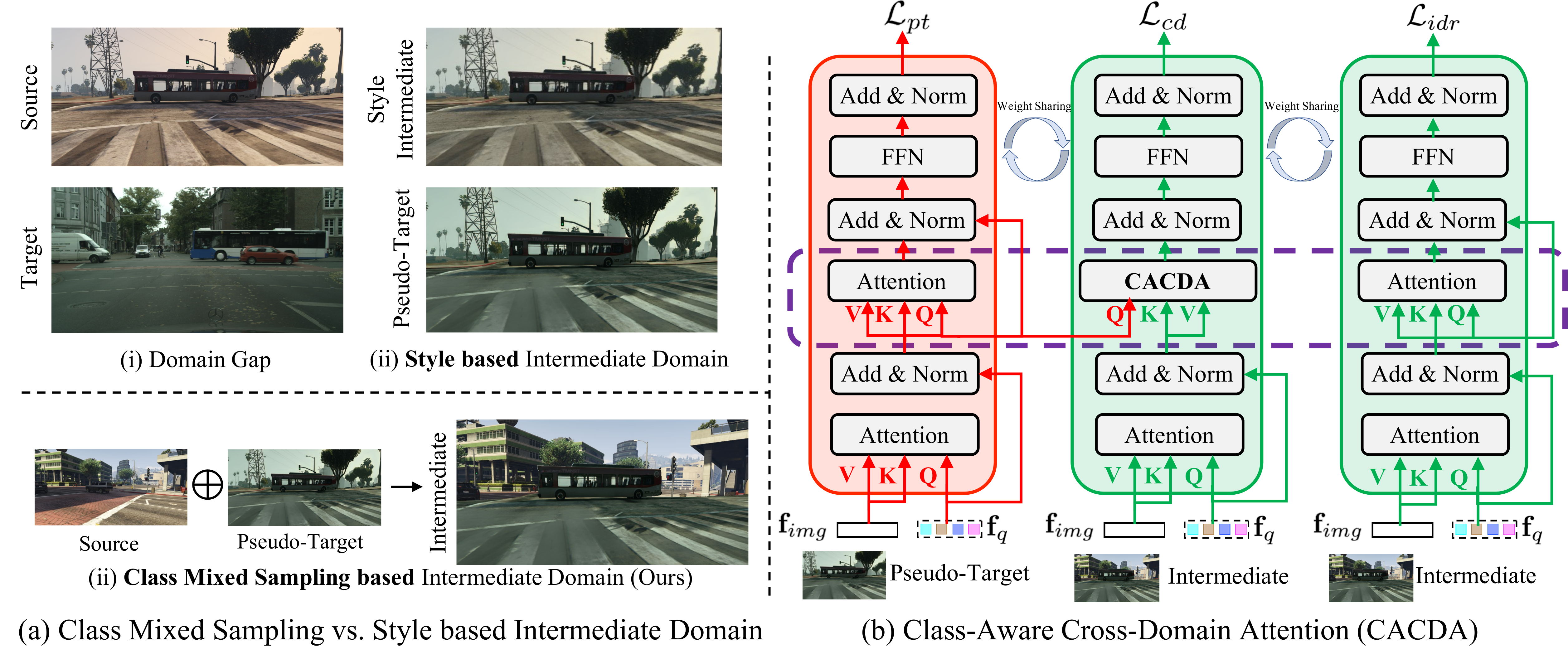}
    \vspace{-20pt}
    \caption{\textbf{Class Mixed Sampling based Intermediate Domain and Cross-Domain Transformers.} (a) (i) shows that the domain gap originates from both the \emph{style} (\eg, the color and texture are different) and \emph{spatial structure} level (\eg, the bus is always located between the crowded city street buildings in the target domain instead of the open field far from buildings in the source domain). (ii) indicates that the style based intermediate domain methods only focus on the \emph{style} differences, \eg, the bus texture is changed, however, is still located in open fields far from buildings as in source domain. (iii) proves that our mixed sampling based intermediate domain can mitigate the domain gap on both the \emph{style} and \emph{spatial structure} level, \eg, the bus is located in the crowded city street buildings in the intermediate domain image, similar to the scene in the target domain image. (b) describes the class-aware cross-domain attention mechanism (cf. purple dashed box) in transformer decoder $\cT$.}
    \vspace{-10pt}
    \label{fig:crossformer}
\end{figure*}

While the pseudo target domain mitigates style differences, spatial structure differences can still exist between the domains. In Fig.~\ref{fig:crossformer}\textcolor{red}{a}, we show an example of this phenomenon. Vehicles are often in an open field and wide roads in the source GTA domain, but are surrounded by crowded city street buildings in the Cityscapes target domain.  City planning differences between virtual US cities and real European cities lead to this significant spatial structural difference.  The target domain contains scene layouts that were never seen before in the source/ pseudo-target domain during training, therefore  model performance  can largely decrease if we do not mitigate the spatial structure differences. 

To reduce the spatial structural domain gap between the source and unseen target images, we propose to randomize the spatial structure between source and pseudo-target using class-mixed sampling.  To this end, we propose a class mixed sampling strategy between the source and pseudo-target domain (cf. Fig.~\ref{fig:crossformer}\textcolor{red}{a}). The sampling strategy  randomizes the layouts between the two domains and aims at improving the generalization performance on the unseen target images. This is largely motivated by existing works on domain randomization~\cite{yue2019domain}. Instead of randomizing the textures of the objects, we randomly copy a class in an pseudo-target image to the source to create a new spatial layout. 

\textbf{Class Mixed Sampling for IDR (CIDR).} More formally, given the pseudo-target domain image $\hat{\x}_i^s$, the corresponding semantic label map $\y_i^s$, and the source domain image $\x_j^s$, the mixed sampling mask $\m^s$ is defined as,
\begin{eqnarray}
    \m^s(h, w) = \begin{cases}
    1, \text{if   } \y_i^s(h, w) = \c, \\
    0, \text{otherwise},
    \end{cases} \label{eq:mask}
\end{eqnarray}
where $(h, w)$ represents the (row, column) index, and $\c$ is the sampling class, which is randomly taken from the available classes in $\y_i^s$. Following~\cite{olsson2021classmix,tranheden2021dacs}, half of the available classes in $\y_i^s$ are randomly selected in each training iteration. Then, the intermediate domain sample $\x_i^m$ and the corresponding semantic label map $\y_i^m$ are written as,
\begin{eqnarray}
    \x_i^m = \m^s \odot \hat{\x}_i^s + (1-\m^s) \odot \x_j^s, \\
    \y_i^m = \m^s \odot \y_i^s + (1-\m^s) \odot \tilde{\y}_j^s,
\end{eqnarray}
where the pseudo-label $\tilde{\y}_j^s$ corresponding to $\x_j^s$, instead of the ground truth label $\y_j^s$, is used to prevent the model from overfitting to the source domain. Then the intermediate domain randomization training loss is written as, $\cL_{idr} = CE(\cF_\theta(\x_i^m), \y_i^m)$. 

\subsection{Class-Aware Cross-Domain Transformers for Domain Invariant Features}\label{sec:mask_aware_trans}
Both the style alignment and the spatial structural alignment are based on the input space. To perform well in the target domain, it is also critical to learn a feature extractor that can provide domain-invariant knowledge.
Existing methods focus on employing the local consistency, \eg, pixel-wise L1 loss in~\cite{yue2019domain} and the local patch-wise prototypical matching in~\cite{wu2021style} between different domains to learn domain-invariant knowledge.
However, as noticed in recent works~\cite{xie2021segformer,strudel2021segmenter}, the local knowledge mining operations are biased towards the local interactions, \eg, the boundary and shape of road and building, leading to insufficient processing of global image context and sub-optimal adaptation. In contrast, without the built-in inductive prior, the transformers by design can capture the global context and interactions between different scene elements, \eg, the scene structure layout of road, sidewalk and person. To this end, we propose to incorporate the class-aware cross-domain transformers into the IDR to extract the \emph{global} domain-invariant knowledge between the intermediate domain image and the pseudo-target domain image.

\textbf{Cross-Domain Transformers.} Following~\cite{cheng2021mask2former,cheng2021maskformer}, our semantic segmentation network $\cF_\theta, \cF_{\theta^\prime}$ is typically composed of the backbone $\cB$, pixel decoder $\cP$ and transformer decoder $\cT$ (cf. Fig.~\ref{fig:framework}). The backbone is used to extract the image feature $\f_{img}$. The pixel decoder gradually up-scales and maps the image feature to the per-pixel embedding, $\e_{pixel}\in \mathbb{R}^{C_\e\times H \times W}$ (cf. Fig.~\ref{fig:framework}\textcolor{red}{b}). The transformer decoder $\cT$ attends to the image feature $\f_{img}$ (cf. Fig.~\ref{fig:framework}\textcolor{red}{b}) and the learnable class query features $\f_{q}$ (cf. Fig.~\ref{fig:framework}\textcolor{red}{b}) to generate the per-class embedding $\e_{class}\in \mathbb{R}^{C_\e\times N}$ of dimension $C_\e$ (cf. Fig.~\ref{fig:framework}\textcolor{red}{b}). The final prediction of $\cF_\theta, \cF_{\theta^\prime}$ is produced by the product $\e_{class}^{T}\times \e_{pixel}$ (cf. Fig.~\ref{fig:framework}\textcolor{red}{b}). The standard self-attention in transformer decoder $\cT$ computes, 
\begin{eqnarray}
    Attention(\Q, \K, \V) = softmax(\Q\K^T)\V,
\end{eqnarray}
where $\Q, \K, \V$ are the query, key and value vectors linearly projected from the input of the self-attention module (cf. Fig.~\ref{fig:crossformer}\textcolor{red}{b}). On this basis, the cross-domain attention is derived by incorporating the query vectors $\Q_{pt}$ from the pseudo-target domain image $\hat{\x}_i^s$ and the key, value vectors $\K_{m}, \V_{m}$ from the intermediate domain $\x_i^m$ into the self-attention module (cf. Fig.~\ref{fig:crossformer}\textcolor{red}{b}), formulated as,
\begin{eqnarray}
    CrossAttention(\Q_{pt}, \K_{m}, \V_{m}) =& \nonumber\\ &\!\!\!\!\!\!\!\!\!\!\!\!\!\!\!\!\!\!\!\!\!\!\!\!\!\!\!\!softmax(\Q_{pt}\K_{m}^T)\V_{m}.\label{eq:cross_atten}
\end{eqnarray}

\textbf{Class-Aware Cross-Domain Transformers.} The area corresponding to the sampling class $\c$ (cf. Eq.~\eqref{eq:mask} and `road, bus' in Fig.~\ref{fig:crossformer}\textcolor{red}{a}) in the intermediate domain image is taken from pseudo-target domain image, sharing the same style and spatial structure knowledge. To promote the domain-invariant knowledge excavation, based on the cross-attention in Eq.~\eqref{eq:cross_atten}, the class-aware cross-domain attention (CACDA) is developed (cf. Fig.~\ref{fig:crossformer}\textcolor{red}{b}) to attend in non-sampled classes regions (cf. `sky, tree, building' in Fig.~\ref{fig:crossformer}\textcolor{red}{a}) where style and scene structure are different, written as,
\begin{eqnarray}
    ClassCrossAtten(\Q_{pt}, \K_{m}, \V_{m}, \M_{c}) = \nonumber\\ softmax(\M_{c} + \Q_{pt}\K_{m}^T)\V_{m}, \label{eq:class_cross_attn}
\end{eqnarray}
where $\M_{c}\in \mathbb{R}^{N\times N}$ is the class-modulating matrix to tailor cross-attention operation, written as,
\begin{eqnarray}
    \M_{c} (x, y) = \begin{cases}
    0, \text{    if   } x \neq \c \text{  and } y \neq \c, \\
    -\infty, \text{    otherwise}.
    \end{cases}
\end{eqnarray}

Then, integrating the cross-domain transformers or the class-aware cross-domain transformers into the intermediate domain randomization training leads to the training loss, $\cL_{cd} = CE(\cF_\theta(\x_i^m|\hat{\x}_i^s), \y_i^m)$. $\cF_\theta(\x_i^m|\hat{\x}_i^s)$ represents that the transformer decoder $\cT$ in $\cF_\theta$ utilizes the cross-attention mechanism in Eq.~\eqref{eq:cross_atten} or the class-aware cross-attention mechanism in Eq.~\eqref{eq:class_cross_attn}. 

\subsection{Joint Training}
With the above pseudo-target domain, class mixed sampling based IDR, class-aware cross-domain transformers and the corresponding training losses, the total loss is,
\begin{eqnarray}
    \cL_{total} = \cL_{pt} + \cL_{idr} + \lambda\cL_{cd},
\end{eqnarray}
where $\lambda$ is the hyper-parameter to balance the cross-domain transformer loss and other loss terms, which is set as 0.01 in our work. With the pseudo-label based self-training strategy in Sec.~\ref{sec:train_strategy}, the model $\cF_{\theta}, \cF_{\theta^\prime}$ are trained end-to-end with the loss $\cL_{total}$. Our proposed method for OSUDA is easy to be applied to OSDG, by replacing the one-shot target image $\x^t$ with the target-relevant image $\x^r$.

%%%%%%%%%%%%%%%%%%%%%%%%%%%%%%%%%%%%%%%%%%%%%
\section{Experiments} 
In order to construct the pseudo-target domain, we stylize the source images with the one-shot target image using the MUNIT method~\cite{huang2018multimodal} and the frequency based Fourier Transform~\cite{yang2020fda} for OSUDA and OSDG, respectively. The perceptual loss weight in~\cite{huang2018multimodal} is set as 2.0. $\beta$ in~\cite{yang2020fda} is set as 0.05. The involved datasets description on, Cityscapes~\cite{cordts2016cityscapes}, GTA~\cite{richter2016playing}, SYNTHIA~\cite{ros2016synthia}, Foggy Cityscapes~\cite{sakaridis2018semantic,sakaridis2018model}, and training details are put in the supplementary.

\subsection{One-Shot Unsupervised Domain Adaptation}

\begin{table*}[t]
    \centering
    \resizebox{\linewidth}{!}{
    \begin{tabular}{c|c|cccccccccccccc|cc}
    \toprule
    Method
    & \rotatebox{60}{Source} & \rotatebox{60}{AdaNet~\cite{Tsai_adaptseg_2018}} & \rotatebox{60}{CLAN~\cite{luo2019taking}} & \rotatebox{60}{AENT~\cite{vu2019advent}} & \rotatebox{60}{CBST~\cite{zou2018unsupervised}} & \rotatebox{60}{CGAN~\cite{CycleGAN2017}} & \rotatebox{60}{OST~\cite{benaim2018one}}  & \rotatebox{60}{FSDR~\cite{huang2021fsdr}} & \rotatebox{60}{DRPC\cite{yue2019domain}} & \rotatebox{60}{SADG\cite{peng2022semantic}} & \rotatebox{60}{RoNet~\cite{choi2021robustnet}} & \rotatebox{60}{WDGE~\cite{kim2021wedge}} & \rotatebox{60}{IBN~\cite{pan2018IBN-Net}} & \rotatebox{60}{ASM~\cite{luo2020adversarial}} & \rotatebox{60}{SPPM\cite{wu2021style}} & \rotatebox{60}{Ours(R)} & \rotatebox{60}{Ours(M)} \\
    \midrule
    Num$^\#$ & 0 & 1 & 1 & 1& 1& 1 & 1  & $\geq$15 & $\geq$15 & 0 &  0 &  1000 & 0 & $\geq$10 & 1 & 1 & 1 \\ 
    \midrule
    \multicolumn{18}{c}{GTA$\rightarrow$ Cityscapes} \\
    \midrule
    mIoU & 36.6 & 35.2 & 37.7 & 36.1 & 37.1 & 39.6 & 42.3  & 44.8 & 42.5 & 45.33 & 42.87 & 43.60 & 37.42 & 44.5 & 42.8 & \textbf{49.46} & \textbf{55.37} \\
    \midrule
    \multicolumn{18}{c}{SYNTHIA$\rightarrow$ Cityscapes} \\
    \midrule
    mIoU$^*$ & 33.65  & 39.1 & 40.4 & 39.9 & 38.5 & 42.1 & 42.8  & 47.3 & - & - & - & - & - & 40.7 & 47.3 & \textbf{51.72} & \textbf{57.80} \\
    mIoU & 29.45 & - & - & - & - & - & -  & 40.8 & 37.6 & 40.87 & 37.21 & 40.31 & 34.18 & 34.6 & 41.4 & \textbf{44.99} & \textbf{50.99} \\
    \bottomrule
    \end{tabular}
    }
    \caption{\textbf{OSUDA Performance.} Num$^{\#}$ is the number of \textbf{real} images used for training, which are from the target domain or other auxiliary datasets, \eg, ImageNet. Baselines and ours (R) adopt ResNet-101 backbone. Ours (M) utilizes MiT-B5 backbone. mIoU$^*$ represents the 13 classes mIoU performance when removing the ``wall, fence, pole" classes, as the common practice in~\cite{Tsai_adaptseg_2018,tranheden2021dacs,vu2019advent}.}
    \vspace{-10pt}
    \label{tab:OSUDA}
\end{table*}

\begin{figure*}
    \centering
    \includegraphics[width=\linewidth]{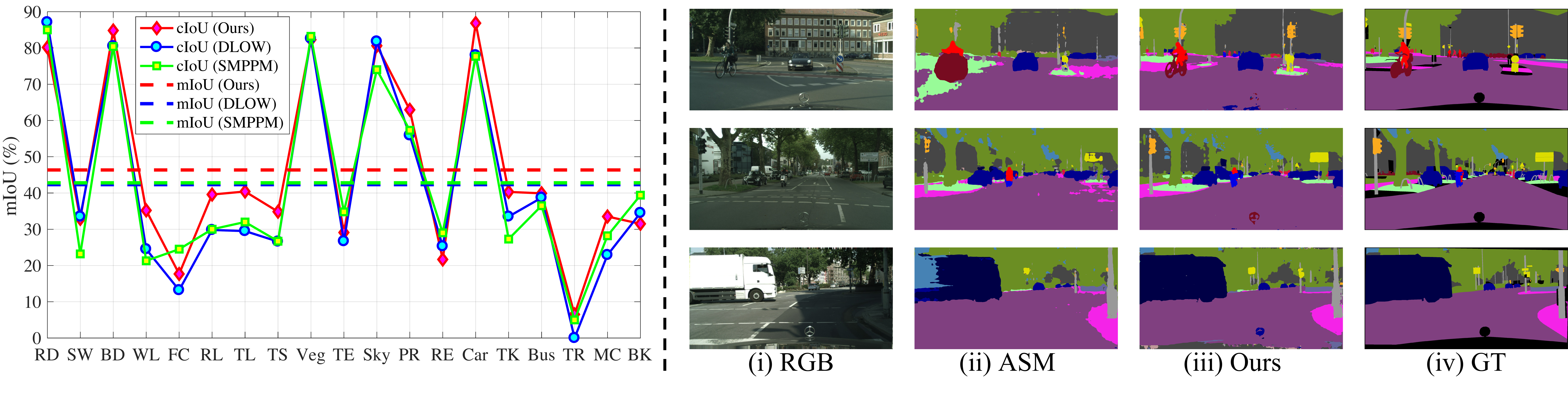}
    \vspace{-20pt}
    \caption{\textbf{Left:} Our class mixed sampling based IDR (46.36 mIoU) \vs other IDR methods, DLOW~\cite{gong2019dlow} (42.3 mIoU) and SMPPM~\cite{wu2021style} (42.8 mIoU), with ResNet-101 backbone under GTA$\rightarrow$Cityscapes. cIoU (solid line) is the each single class IoU performance, while the mIoU (dashed line) is the mean IoU performance over cIoU. \textbf{Right:} Qualitative semantic segmentation results comparison on the target domain between our method and the baseline method ASM~\cite{luo2020adversarial}, where GT is the ground truth semantic label map.}
    \label{fig:idr_quali}
    \vspace{-10pt}
\end{figure*}

\textbf{Comparison to other SOTA methods.} To demonstrate the effectiveness of our method for OSUDA, we compare our methods with other SOTA methods on the popular benchmarks, GTA, SYNTHIA$\rightarrow$Cityscapes, respectively. The baseline methods include 1) the OSUDA methods, ASM~\cite{luo2020adversarial} and SPPM~\cite{wu2021style}; 2) the UDA methods, AdaNet~\cite{Tsai_adaptseg_2018}, CLAN~\cite{luo2019taking}, AENT~\cite{vu2019advent}, CBST~\cite{zou2018unsupervised}, CGAN~\cite{CycleGAN2017} and OST~\cite{benaim2018one}; 3) the DG/DR methods, FSDR~\cite{huang2021fsdr}, DRPC~\cite{yue2019domain}, RoNet~\cite{choi2021robustnet}, IBN~\cite{pan2018IBN-Net}, WDGE~\cite{kim2021wedge} and SADG~\cite{peng2022semantic}. From Table~\ref{tab:OSUDA}, it is shown that our proposed method outperforms the previous SOTA OSUDA methods with the same ResNet-101~\cite{he2016deep} backbone, 49.46\% \vs 44.5\%, 42.8\% for GTA, and 44.99\% \vs 34.6\%, 41.4\% for SYNTHIA. By replacing the ResNet-101 with the transformer based backbone MiT-B5~\cite{xie2021segformer}, the performance can be further improved to 55.37\% and 50.99\%, benefiting from the stronger knowledge extraction ability of transformers compared to convolutional neural networks (CNNs). Besides, the previous OSUDA method, ASM~\cite{luo2020adversarial}, and DG/DR methods, FSDR~\cite{huang2021fsdr}, WEDGE~\cite{kim2021wedge} and DRPC~\cite{yue2019domain}, typically utilize a large number of auxiliary real images (\eg, $\geq$1000 for WEDGE) of diverse styles from ImageNet~\cite{deng2009imagenet} and other datasets for training to improve the sim-to-real domain generalization ability. Nevertheless, our method does not require these auxiliary real images, while still strongly outperforming the aforementioned methods. In Fig.~\ref{fig:idr_quali}, we show the qualitative comparison results of our proposed method and other methods.

\begin{figure*}
    \centering
    \includegraphics[width=0.9\linewidth]{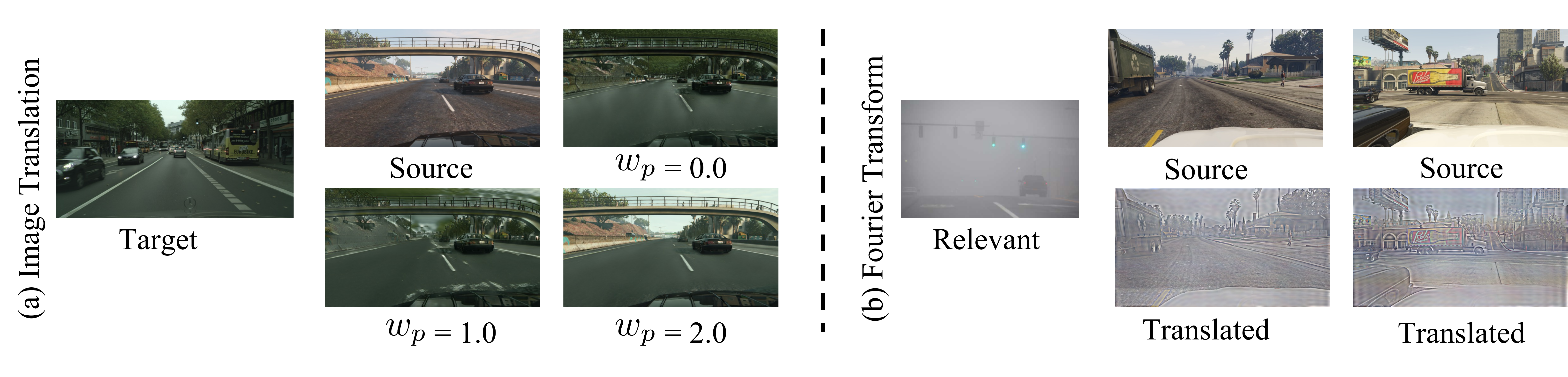}
    \vspace{-15pt}
    \caption{\textbf{Pseudo-Target Domain Construction}. (a) is the deep learning based image translation method. (b) is the non-learning based Fourier Transform method. It proves the effectiveness and flexibility of our framework for utilizing different types of pseudo-target domain construction methods, learning/non-learning based approach.}
    \label{fig:pt_example}
    \vspace{-5pt}
    % \vspace{-10pt}
\end{figure*}

\textbf{Pseudo-target domain construction.} In order to prove the effectiveness of our pseudo-target domain construction for transferring the target domain style and preserving the content of the source domain, we show the qualitative pseudo-target domain construction results in Fig.~\ref{fig:pt_example}. It is shown that our proposed framework is flexible, and different types of pseudo-target domain construction methods, learning/non-learning based methods, can be used.
\emph{Learning based approach:} In order to construct the pseudo-target domain, for OSUDA experiments, we stylize the images with the one-shot target image by using the deep learning based MUNIT method~\cite{huang2018multimodal}, setting the perceptual loss weight $w_p$ as 2.0. As shown in Fig.~\ref{fig:pt_example}\textcolor{red}{a}, by adding the higher weight $w_p$ to the perceptual loss, the example guided image translation method can translate the source-domain image to the pseudo-target domain image, without overfitting to the one-shot target image. In other words, the style of the target image is applied to the source domain image, while the content from the source domain image is preserved. It provides a simple yet effective strategy for the one-shot/ few-shot image translation, \ie, adding higher weighted perceptual loss. \emph{Non-learning based approach:} Moreover, in order to prove the flexibility of our proposed whole framework for using different ways of pseudo-target domain construction, we adopt the frequency based Fourier Transform method~\cite{yang2020fda} for the OSDG experiments, which is non-learning based approach. The advantage of Fourier Transform is that it does not require the training process of the deep learning based approach (\eg, the image translation method in Fig.~\ref{fig:pt_example}\textcolor{red}{a}), and is easy to be used and applied. The disadvantage is that it introduces more artifacts than the deep learning based image translation method, which is also shown in \cite{yang2020fda}. However, as shown in Fig.~\ref{fig:pt_example}\textcolor{red}{b}, though the artifacts exist, the style of the target relevant image is still transferred to the source domain and the content in the source domain image is still preserved. The OSDG experiments in Table~\ref{tab:OSDG} prove that, our framework can effectively realize the one-shot sim-to-real domain generalization with the Fourier Transform based image, verifying our framework is compatible with different learning/ non-learning based pseudo-target domain construction methods.

\begin{table}
    \centering
    \resizebox{0.8\linewidth}{!}{
    \begin{tabular}{c|cc|cc}
    \toprule
        \multirow{2}{*}{Setting}& \multicolumn{2}{c}{Res101} & \multicolumn{2}{c}{Mit-B5}\\
        \cmidrule{2-5}
        &DAFormer & Ours & DAFormer & Ours\\
        \midrule
        \midrule
        G$\rightarrow$C & 48.53 & \textbf{49.46} &54.01 & \textbf{55.37}\\
        S$\rightarrow$C & 43.79 & \textbf{44.99} & 48.90 & \textbf{50.99} \\
        \midrule
        G$\rightarrow$FC & 38.57 & \textbf{39.05} & 47.43 & \textbf{49.04}\\
        S$\rightarrow$FC & 37.75 & \textbf{41.09} & 43.36 & \textbf{44.28}\\
        \bottomrule
    \end{tabular}
    }
    \vspace{-5pt}
    \caption{\textbf{Comparison to DAFormer~\cite{hoyer2022daformer} for OSUDA and OSDG.} \cite{hoyer2022daformer} is combined with PT and CIDR for fair comparison.}
    \label{tab:daformer_comp}
    \vspace{-15pt}
    % \vspace{3pt}
\end{table}

\begin{table}
    \centering
    \resizebox{0.8\linewidth}{!}{
    \begin{tabular}{cccc|c}
    \toprule
    PT & CIDR & CrossAtten & Class-Aware & mIoU \\
        \midrule
        \midrule
    &&&&42.59\\
    \checkmark& & & & 46.12\\
    \checkmark& \checkmark & & & 53.45\\
    \checkmark& \checkmark & \checkmark & & 54.62 \\
    \checkmark& \checkmark & \checkmark & \checkmark & \textbf{55.37}\\
        \bottomrule
    \end{tabular}
    }
    \caption{\textbf{Ablation Study.} Experiments are conducted under OSUDA setting, GTA $\rightarrow$ Cityscapes, with the backbone as MiT-B5.}
    \label{tab:ablation}
    \vspace{-10pt}
    % \vspace{3pt}
\end{table}

\begin{table*}
    \centering
    \resizebox{\linewidth}{!}{
    \begin{tabular}{c|c|cccccccc|cc}
    \toprule
    Method
    & Source & BDL~\cite{li2019bidirectional} & FDA~\cite{yang2020fda} & GASF\cite{Kundu_2021_ICCV} & BDL~\cite{li2019bidirectional} & CBST~\cite{zou2018unsupervised} & MLSL~\cite{iqbal2020mlsl}  & FAdapt~\cite{iqbal2022fogadapt} & ASM~\cite{luo2020adversarial} & Ours(R) & Ours(M) \\
    \midrule
    Num$^\#$ & 0 & 0 & 0 & 0 & 2975 & 2975 & 2975 & 2975  & 0 & 0 & 0 \\ 
    \midrule
    \multicolumn{12}{c}{GTA$\rightarrow$Foggy Cityscapes} \\
    \midrule
    mIoU & 33.4 & 30.3 & 35.3  & 38.3 & 36.3 &  37.7  & 39.1 & 41.0 & 35.99 & \textbf{39.05} & \textbf{49.04} \\
    \midrule
    \multicolumn{12}{c}{SYNTHIA$\rightarrow$Foggy Cityscapes} \\
    \midrule
    mIoU$^*$ & 24.4 & - & - & - & 38.1 & 38.4 & 40.8 & 41.4 & 33.12 & \textbf{47.28} & \textbf{50.17} \\
    mIoU & 20.9 & - & - & - & 32.3 & 33.3 & 35.9 & 36.4 & 28.37 & \textbf{41.09} & \textbf{44.28} \\
    \bottomrule
    \end{tabular}
    }
    \caption{\textbf{OSDG Performance.} Num$^{\#}$ represents the number of \textbf{Foggy Cityscapes} (target domain) images used for training. Baselines and ours (R) adopt ResNet-101 backbone. Ours (M) utilizes MiT-B5 backbone. mIoU$^*$ represents the 13 classes mIoU performance when removing the ``wall, fence, pole" classes, as the common practice in~\cite{Tsai_adaptseg_2018,tranheden2021dacs,vu2019advent}.}
    \label{tab:OSDG}
    \vspace{-5pt}
\end{table*}

\begin{figure*}
    \centering
    \includegraphics[width=\linewidth]{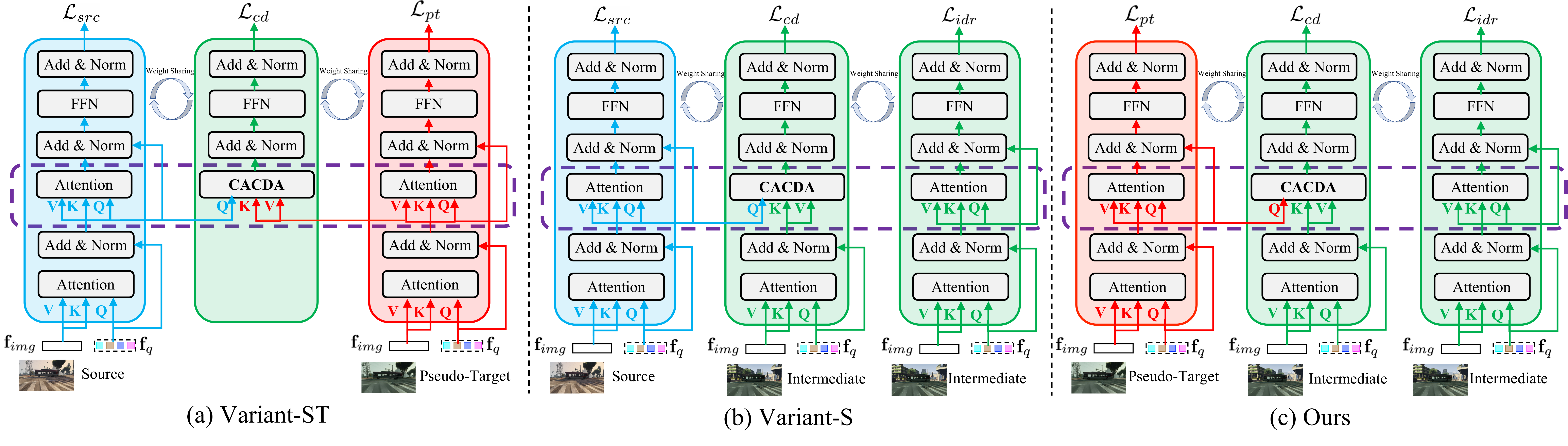}
    \vspace{-20pt}
    \caption{\textbf{Different Cross-Domain Transformer Variants.} (a) is the variant-ST, where the supervised training is conducted on the source domain and pseudo-target domain, and then the cross-attention is utilized between the source and pseudo-target domain. (b) is the vairiant-S, where the supervised training is conducted on the source domain, and then the cross-attention is utilized between the source and intermediate domain. (c) is our cross-domain transformer module structure in Sec.~\ref{sec:mask_aware_trans}, where the supervised training is conducted on the pseudo-target domain, and then the cross-attention is utilized between the pseudo-target domain and intermediate domain. $\cL_{src}$ is the supervised cross-entropy training loss on the source domain.}
    \label{fig:cross_att_supp}
    \vspace{-10pt}
\end{figure*}

\begin{table}
    \centering
    \begin{tabular}{c|ccc}
    \toprule
         Structure& Variant-ST & Variant-S & Ours\\
         \midrule
         mIoU & 52.54 & 54.05 & \textbf{55.37}\\
         \bottomrule
    \end{tabular}
    \caption{\textbf{Different Cross-Domain Transformer Variants.} The experiments are conducted under GTA$\rightarrow$Cityscapes for OSUDA, with the MiT-B5 backbone.}
    \vspace{-10pt}
    \label{tab:cross_att_supp}
\end{table}

\textbf{Ablation Study.} We investigate the individual contributions of different modules towards the overall performance quantitatively in Table~\ref{tab:ablation}. The experiments are conducted under GTA$\rightarrow$Cityscapes, with the MiT-B5 backbone. It is shown that all modules, including pseudo-target domain, class mixed sampling based IDR, and class-aware cross-domain transformers, are able to contribute to the final performance and help boost the adaptation ability of the model. The pseudo-target domain improves the performance from 42.59\% to 46.12\%, proving the effectiveness of the \emph{style} knowledge from the one-shot target domain. The class mixed sampling based intermediate domain randomization further narrows the domain gap on \emph{style} and \emph{spatial structure}, reaching 53.45\%. Moreover, the class-aware cross-domain transformers extract the global domain-invariant knowledge, leading to 55.37\% performance and boosting other local knowledge extraction methods such as DRPC~\cite{yue2019domain} and FSDR~\cite{huang2021fsdr}. Besides, the class-aware cross-domain attention performs better than pure cross-domain attention, 55.37\% \vs 54.62\%. This is because the class-aware cross-domain attention attends to the non-sampled class regions, where the style and the spatial structure are different. Thus, it encourages and promotes the domain-invariant knowledge extraction. In Table~\ref{tab:daformer_comp}, we compare to the SOTA transformer based domain adaptation method, DAFormer~\cite{hoyer2022daformer}, and it is observed that our method outperforms DAFormer under various settings.

\textbf{Class Mixed Sampling based IDR \vs other IDR methods.} As shown in Fig.~\ref{fig:idr_quali}, the image-level and feature-level stylization based intermediate domain methods achieve similar performance, 42.3\% and 42.8\% under GTA$\rightarrow$Cityscapes. Nevertheless, our class mixed sampling based IDR method reaches 46.36\%. That is because, compared to the previous IDR methods such as DLOW~\cite{gong2019dlow,gong2021dlow} and SMPPM~\cite{wu2021style} that only focus on \emph{style}, our class mixed sampling based IDR can bridge the domain gap on both \emph{style} and \emph{spatial structure} level (cf. Fig.~\ref{fig:crossformer}\textcolor{red}{a}). 

\textbf{Comparison to Cross-Domain Transformer Variants.} In Sec.~\ref{sec:mask_aware_trans}, we propose the class-aware cross-domain transformers, where the class-aware cross-domain attention is conducted between the pseudo-target domain and the intermediate domain. In order to demonstrate the effectiveness of our proposed cross-domain transformer module structure in Sec.~\ref{sec:mask_aware_trans}, we here compare it with other cross-domain variants, where different cross-attention strategies between source domain, intermediate domain and pseudo-target domain are conducted as shown in Fig.~\ref{fig:cross_att_supp}. In Table~\ref{tab:cross_att_supp}, we show the quantitative comparison results between our cross-domain transformer structure and other cross-domain transformer variants, variant-ST (Fig.~\ref{fig:cross_att_supp}\textcolor{red}{a}) and variant-S (Fig.~\ref{fig:cross_att_supp}\textcolor{red}{b}). It is observed that both our structure in Sec.~\ref{sec:mask_aware_trans} and variant-S outperform the variant-ST, 55.37\%, 54.05\% \vs 52.54\%. It proves the effectiveness and necessity of conducting the cross-domain attention on the intermediate domain, which bridges the source and the target domain and eases the domain-invariant knowledge extraction through the cross-domain attention. Besides, our structure in Sec.~\ref{sec:mask_aware_trans} performs better than the variant-S, 55.37\% \vs 54.05\%. It is because that the pseudo-target domain is closer to the target domain compared to the source domain, and eases the adaptation to the target domain, verifying the effectiveness of the pseudo-target domain construction.

\subsection{One-Shot Sim-to-Real Domain Generalization}
Similar to the experimental study on OSUDA, we compare our proposed method with other SOTA methods under OSDG setting in Table~\ref{tab:OSDG}. Remarkably, our proposed method not only performs better than the SOTA OSUDA method ASM, but also performs at par or even outperforms the SOTA UDA methods under GTA, SYNTHIA $\rightarrow$ Foggy Cityscapes. It is notable that the UDA methods utilize the whole 2975 target domain foggy cityscapes images for training, while our method only exploits one relevant foggy image from the website~\cite{sakaridis2018semantic}.

\section{Conclusion}
In this paper, we aim at addressing the one-shot unsupervised sim-to-real domain adaptation (OSUDA) and generalization (OSDG) problem. To this end, we propose a transformer based approach, where 1) the pseudo-target domain and class-mixed sampling based intermediate domain is proposed to bridge the domain gap on both the style and spatial structure level; 2) the class-aware cross-domain transformers further extract the domain-invariant knowledge, facilitating the adaptation to the target domain. Extensive experiments on different benchmarks prove the effectiveness of our proposed method for OSUDA and OSDG, improving the SOTA performance by a large margin.

%%%%%%%%% REFERENCES
{\small
\bibliographystyle{ieee_fullname}
\bibliography{egbib}

\begin{thebibliography}{10}\itemsep=-1pt

\bibitem{benaim2018one}
Sagie Benaim and Lior Wolf.
\newblock One-shot unsupervised cross domain translation.
\newblock In {\em NeurIPS}, 2018.

\bibitem{bousmalis2018using}
Konstantinos Bousmalis, Alex Irpan, Paul Wohlhart, Yunfei Bai, Matthew Kelcey,
  Mrinal Kalakrishnan, Laura Downs, Julian Ibarz, Peter Pastor, Kurt Konolige,
  et~al.
\newblock Using simulation and domain adaptation to improve efficiency of deep
  robotic grasping.
\newblock In {\em ICRA}, 2018.

\bibitem{carion2020end}
Nicolas Carion, Francisco Massa, Gabriel Synnaeve, Nicolas Usunier, Alexander
  Kirillov, and Sergey Zagoruyko.
\newblock End-to-end object detection with transformers.
\newblock In {\em ECCV}, 2020.

\bibitem{chen2017deeplab}
Liang-Chieh Chen, George Papandreou, Iasonas Kokkinos, Kevin Murphy, and Alan~L
  Yuille.
\newblock Deeplab: Semantic image segmentation with deep convolutional nets,
  atrous convolution, and fully connected crfs.
\newblock {\em TPAMI}, 40(4):834--848, 2017.

\bibitem{cheng2021mask2former}
Bowen Cheng, Ishan Misra, Alexander~G. Schwing, Alexander Kirillov, and Rohit
  Girdhar.
\newblock Masked-attention mask transformer for universal image segmentation.
\newblock In {\em CVPR}, 2022.

\bibitem{cheng2021maskformer}
Bowen Cheng, Alexander~G. Schwing, and Alexander Kirillov.
\newblock Per-pixel classification is not all you need for semantic
  segmentation.
\newblock In {\em NeurIPS}, 2021.

\bibitem{choi2021robustnet}
Sungha Choi, Sanghun Jung, Huiwon Yun, Joanne~T Kim, Seungryong Kim, and Jaegul
  Choo.
\newblock Robustnet: Improving domain generalization in urban-scene
  segmentation via instance selective whitening.
\newblock In {\em CVPR}, 2021.

\bibitem{cordts2016cityscapes}
Marius Cordts, Mohamed Omran, Sebastian Ramos, Timo Rehfeld, Markus Enzweiler,
  Rodrigo Benenson, Uwe Franke, Stefan Roth, and Bernt Schiele.
\newblock The cityscapes dataset for semantic urban scene understanding.
\newblock In {\em CVPR}, 2016.

\bibitem{deng2009imagenet}
Jia Deng, Wei Dong, Richard Socher, Li-Jia Li, Kai Li, and Li Fei-Fei.
\newblock Imagenet: A large-scale hierarchical image database.
\newblock In {\em CVPR}, 2009.

\bibitem{devaranjan2020meta}
Jeevan Devaranjan, Amlan Kar, and Sanja Fidler.
\newblock Meta-sim2: Unsupervised learning of scene structure for synthetic
  data generation.
\newblock In {\em ECCV}, 2020.

\bibitem{dosovitskiy2020vit}
Alexey Dosovitskiy, Lucas Beyer, Alexander Kolesnikov, Dirk Weissenborn,
  Xiaohua Zhai, Thomas Unterthiner, Mostafa Dehghani, Matthias Minderer, Georg
  Heigold, Sylvain Gelly, Jakob Uszkoreit, and Neil Houlsby.
\newblock An image is worth 16x16 words: Transformers for image recognition at
  scale.
\newblock In {\em ICLR}, 2021.

\bibitem{dosovitskiy2017carla}
Alexey Dosovitskiy, German Ros, Felipe Codevilla, Antonio Lopez, and Vladlen
  Koltun.
\newblock Carla: An open urban driving simulator.
\newblock In {\em CORL}, 2017.

\bibitem{fang2018multi}
Kuan Fang, Yunfei Bai, Stefan Hinterstoisser, Silvio Savarese, and Mrinal
  Kalakrishnan.
\newblock Multi-task domain adaptation for deep learning of instance grasping
  from simulation.
\newblock In {\em ICRA}, 2018.

\bibitem{gogoll2020unsupervised}
Dario Gogoll, Philipp Lottes, Jan Weyler, Nik Petrinic, and Cyrill Stachniss.
\newblock Unsupervised domain adaptation for transferring plant classification
  systems to new field environments, crops, and robots.
\newblock In {\em IROS}, 2020.

\bibitem{gong2021dlow}
Rui Gong, Wen Li, Yuhua Chen, Dengxin Dai, and Luc Van~Gool.
\newblock Dlow: Domain flow and applications.
\newblock {\em IJCV}, 129(10):2865--2888, 2021.

\bibitem{gong2019dlow}
Rui Gong, Wen Li, Yuhua Chen, and Luc~Van Gool.
\newblock Dlow: Domain flow for adaptation and generalization.
\newblock In {\em CVPR}, 2019.

\bibitem{he2016deep}
Kaiming He, Xiangyu Zhang, Shaoqing Ren, and Jian Sun.
\newblock Deep residual learning for image recognition.
\newblock In {\em CVPR}, 2016.

\bibitem{hoffman2018cycada}
Judy Hoffman, Eric Tzeng, Taesung Park, Jun-Yan Zhu, Phillip Isola, Kate
  Saenko, Alexei Efros, and Trevor Darrell.
\newblock Cycada: Cycle-consistent adversarial domain adaptation.
\newblock In {\em ICML}, 2018.

\bibitem{hoyer2022daformer}
Lukas Hoyer, Dengxin Dai, and Luc Van~Gool.
\newblock Daformer: Improving network architectures and training strategies for
  domain-adaptive semantic segmentation.
\newblock In {\em CVPR}, 2022.

\bibitem{huang2021fsdr}
Jiaxing Huang, Dayan Guan, Aoran Xiao, and Shijian Lu.
\newblock Fsdr: Frequency space domain randomization for domain generalization.
\newblock In {\em CVPR}, 2021.

\bibitem{huang2017arbitrary}
Xun Huang and Serge Belongie.
\newblock Arbitrary style transfer in real-time with adaptive instance
  normalization.
\newblock In {\em ICCV}, 2017.

\bibitem{huang2018multimodal}
Xun Huang, Ming-Yu Liu, Serge Belongie, and Jan Kautz.
\newblock Multimodal unsupervised image-to-image translation.
\newblock In {\em ECCV}, 2018.

\bibitem{iqbal2020mlsl}
Javed Iqbal and Mohsen Ali.
\newblock Mlsl: Multi-level self-supervised learning for domain adaptation with
  spatially independent and semantically consistent labeling.
\newblock In {\em WACV}, 2020.

\bibitem{iqbal2022fogadapt}
Javed Iqbal, Rehan Hafiz, and Mohsen Ali.
\newblock Fogadapt: Self-supervised domain adaptation for semantic segmentation
  of foggy images.
\newblock {\em Neurocomputing}, 2022.

\bibitem{johnson2016perceptual}
Justin Johnson, Alexandre Alahi, and Li Fei-Fei.
\newblock Perceptual losses for real-time style transfer and super-resolution.
\newblock In {\em ECCV}, 2016.

\bibitem{kar2019meta}
Amlan Kar, Aayush Prakash, Ming-Yu Liu, Eric Cameracci, Justin Yuan, Matt
  Rusiniak, David Acuna, Antonio Torralba, and Sanja Fidler.
\newblock Meta-sim: Learning to generate synthetic datasets.
\newblock In {\em ICCV}, 2019.

\bibitem{keser2021content}
Mert Keser, Artem Savkin, and Federico Tombari.
\newblock Content disentanglement for semantically consistent synthetic-to-real
  domain adaptation.
\newblock In {\em IROS}, 2021.

\bibitem{kim2021wedge}
Namyup Kim, Taeyoung Son, Cuiling Lan, Wenjun Zeng, and Suha Kwak.
\newblock Wedge: Web-image assisted domain generalization for semantic
  segmentation.
\newblock {\em arXiv preprint arXiv:2109.14196}, 2021.

\bibitem{Kundu_2021_ICCV}
Jogendra~Nath Kundu, Akshay Kulkarni, Amit Singh, Varun Jampani, and
  R.~Venkatesh Babu.
\newblock Generalize then adapt: Source-free domain adaptive semantic
  segmentation.
\newblock In {\em ICCV}, 2021.

\bibitem{li2019bidirectional}
Yunsheng Li, Lu Yuan, and Nuno Vasconcelos.
\newblock Bidirectional learning for domain adaptation of semantic
  segmentation.
\newblock In {\em CVPR}, 2019.

\bibitem{liu2021swin}
Ze Liu, Yutong Lin, Yue Cao, Han Hu, Yixuan Wei, Zheng Zhang, Stephen Lin, and
  Baining Guo.
\newblock Swin transformer: Hierarchical vision transformer using shifted
  windows.
\newblock In {\em ICCV}, 2021.

\bibitem{long2015fully}
Jonathan Long, Evan Shelhamer, and Trevor Darrell.
\newblock Fully convolutional networks for semantic segmentation.
\newblock In {\em CVPR}, 2015.

\bibitem{loshchilov2019decoupled}
Ilya Loshchilov and Frank Hutter.
\newblock Decoupled weight decay regularization.
\newblock In {\em ICLR}, 2019.

\bibitem{luo2020adversarial}
Yawei Luo, Ping Liu, Tao Guan, Junqing Yu, and Yi Yang.
\newblock Adversarial style mining for one-shot unsupervised domain adaptation.
\newblock In {\em NeurIPS}, 2020.

\bibitem{luo2019taking}
Yawei Luo, Liang Zheng, Tao Guan, Junqing Yu, and Yi Yang.
\newblock Taking a closer look at domain shift: Category-level adversaries for
  semantics consistent domain adaptation.
\newblock In {\em CVPR}, 2019.

\bibitem{messikommer2022bridging}
Nico Messikommer, Daniel Gehrig, Mathias Gehrig, and Davide Scaramuzza.
\newblock Bridging the gap between events and frames through unsupervised
  domain adaptation.
\newblock {\em RA-L}, 7(2):3515--3522, 2022.

\bibitem{olsson2021classmix}
Viktor Olsson, Wilhelm Tranheden, Juliano Pinto, and Lennart Svensson.
\newblock Classmix: Segmentation-based data augmentation for semi-supervised
  learning.
\newblock In {\em WACV}, 2021.

\bibitem{palazzo2020domain}
Simone Palazzo, Dario~C Guastella, Luciano Cantelli, Paolo Spadaro, Francesco
  Rundo, Giovanni Muscato, Daniela Giordano, and Concetto Spampinato.
\newblock Domain adaptation for outdoor robot traversability estimation from
  rgb data with safety-preserving loss.
\newblock In {\em IROS}, 2020.

\bibitem{pan2018IBN-Net}
Xingang Pan, Ping Luo, Jianping Shi, and Xiaoou Tang.
\newblock Two at once: Enhancing learning and generalization capacities via
  ibn-net.
\newblock In {\em ECCV}, 2018.

\bibitem{NEURIPS2019_9015}
Adam Paszke, Sam Gross, Francisco Massa, Adam Lerer, James Bradbury, Gregory
  Chanan, Trevor Killeen, Zeming Lin, Natalia Gimelshein, Luca Antiga, Alban
  Desmaison, Andreas Kopf, Edward Yang, Zachary DeVito, Martin Raison, Alykhan
  Tejani, Sasank Chilamkurthy, Benoit Steiner, Lu Fang, Junjie Bai, and Soumith
  Chintala.
\newblock Pytorch: An imperative style, high-performance deep learning library.
\newblock In {\em NeurIPS}, 2019.

\bibitem{peng2022semantic}
Duo Peng, Yinjie Lei, Munawar Hayat, Yulan Guo, and Wen Li.
\newblock Semantic-aware domain generalized segmentation.
\newblock In {\em CVPR}, 2022.

\bibitem{prakash2019structured}
Aayush Prakash, Shaad Boochoon, Mark Brophy, David Acuna, Eric Cameracci,
  Gavriel State, Omer Shapira, and Stan Birchfield.
\newblock Structured domain randomization: Bridging the reality gap by
  context-aware synthetic data.
\newblock In {\em ICRA}, 2019.

\bibitem{richter2016playing}
Stephan~R Richter, Vibhav Vineet, Stefan Roth, and Vladlen Koltun.
\newblock Playing for data: Ground truth from computer games.
\newblock In {\em ECCV}, 2016.

\bibitem{ros2016synthia}
German Ros, Laura Sellart, Joanna Materzynska, David Vazquez, and Antonio~M
  Lopez.
\newblock The synthia dataset: A large collection of synthetic images for
  semantic segmentation of urban scenes.
\newblock In {\em CVPR}, 2016.

\bibitem{sakaridis2018model}
Christos Sakaridis, Dengxin Dai, Simon Hecker, and Luc Van~Gool.
\newblock Model adaptation with synthetic and real data for semantic dense
  foggy scene understanding.
\newblock In {\em ECCV}, 2018.

\bibitem{sakaridis2018semantic}
Christos Sakaridis, Dengxin Dai, and Luc Van~Gool.
\newblock Semantic foggy scene understanding with synthetic data.
\newblock {\em IJCV}, 126(9):973--992, 2018.

\bibitem{sakuma2021geometry}
Hiroki Sakuma and Yoshinori Konishi.
\newblock Geometry-aware unsupervised domain adaptation for stereo matching.
\newblock In {\em ICRA}, 2021.

\bibitem{strudel2021segmenter}
Robin Strudel, Ricardo Garcia, Ivan Laptev, and Cordelia Schmid.
\newblock Segmenter: Transformer for semantic segmentation.
\newblock In {\em ICCV}, 2021.

\bibitem{tarvainen2017mean}
Antti Tarvainen and Harri Valpola.
\newblock Mean teachers are better role models: Weight-averaged consistency
  targets improve semi-supervised deep learning results.
\newblock In {\em NeurIPS}, 2017.

\bibitem{tobin2017domain}
Josh Tobin, Rachel Fong, Alex Ray, Jonas Schneider, Wojciech Zaremba, and
  Pieter Abbeel.
\newblock Domain randomization for transferring deep neural networks from
  simulation to the real world.
\newblock In {\em IROS}, 2017.

\bibitem{tranheden2021dacs}
Wilhelm Tranheden, Viktor Olsson, Juliano Pinto, and Lennart Svensson.
\newblock Dacs: Domain adaptation via cross-domain mixed sampling.
\newblock In {\em WACV}, 2021.

\bibitem{Tsai_adaptseg_2018}
Y.-H. Tsai, W.-C. Hung, S. Schulter, K. Sohn, M.-H. Yang, and M. Chandraker.
\newblock Learning to adapt structured output space for semantic segmentation.
\newblock In {\em CVPR}, 2018.

\bibitem{vaswani2017attention}
Ashish Vaswani, Noam Shazeer, Niki Parmar, Jakob Uszkoreit, Llion Jones,
  Aidan~N Gomez, Lukasz Kaiser, and Illia Polosukhin.
\newblock Attention is all you need.
\newblock In {\em NeurIPS}, 2017.

\bibitem{vu2019advent}
Tuan-Hung Vu, Himalaya Jain, Maxime Bucher, Matthieu Cord, and Patrick
  P{\'e}rez.
\newblock Advent: Adversarial entropy minimization for domain adaptation in
  semantic segmentation.
\newblock In {\em CVPR}, 2019.

\bibitem{wu2021style}
Xinyi Wu, Zhenyao Wu, Yuhang Lu, Lili Ju, and Song Wang.
\newblock Style mixing and patchwise prototypical matching for one-shot
  unsupervised domain adaptive semantic segmentation.
\newblock In {\em AAAI}, 2022.

\bibitem{wulfmeier2017addressing}
Markus Wulfmeier, Alex Bewley, and Ingmar Posner.
\newblock Addressing appearance change in outdoor robotics with adversarial
  domain adaptation.
\newblock In {\em IROS}, 2017.

\bibitem{wulfmeier2018incremental}
Markus Wulfmeier, Alex Bewley, and Ingmar Posner.
\newblock Incremental adversarial domain adaptation for continually changing
  environments.
\newblock In {\em ICRA}, 2018.

\bibitem{xie2021segformer}
Enze Xie, Wenhai Wang, Zhiding Yu, Anima Anandkumar, Jose~M Alvarez, and Ping
  Luo.
\newblock Segformer: Simple and efficient design for semantic segmentation with
  transformers.
\newblock In {\em NeurIPS}, 2021.

\bibitem{xu2022cdtrans}
Tongkun Xu, Weihua Chen, Pichao Wang, Fan Wang, Hao Li, and Rong Jin.
\newblock Cdtrans: Cross-domain transformer for unsupervised domain adaptation.
\newblock In {\em ICLR}, 2022.

\bibitem{yang2020fda}
Yanchao Yang and Stefano Soatto.
\newblock Fda: Fourier domain adaptation for semantic segmentation.
\newblock In {\em CVPR}, 2020.

\bibitem{yue2019domain}
Xiangyu Yue, Yang Zhang, Sicheng Zhao, Alberto Sangiovanni-Vincentelli, Kurt
  Keutzer, and Boqing Gong.
\newblock Domain randomization and pyramid consistency: Simulation-to-real
  generalization without accessing target domain data.
\newblock In {\em ICCV}, 2019.

\bibitem{yun2021target}
Woo-han Yun, ByungOk Han, Jaeyeon Lee, Jaehong Kim, and Junmo Kim.
\newblock Target-style-aware unsupervised domain adaptation for object
  detection.
\newblock {\em RA-L}, 6(2):3825--3832, 2021.

\bibitem{zhang2019vr}
Jingwei Zhang, Lei Tai, Peng Yun, Yufeng Xiong, Ming Liu, Joschka Boedecker,
  and Wolfram Burgard.
\newblock Vr-goggles for robots: Real-to-sim domain adaptation for visual
  control.
\newblock {\em RA-L}, 4(2):1148--1155, 2019.

\bibitem{zheng2021rethinking}
Sixiao Zheng, Jiachen Lu, Hengshuang Zhao, Xiatian Zhu, Zekun Luo, Yabiao Wang,
  Yanwei Fu, Jianfeng Feng, Tao Xiang, Philip~HS Torr, et~al.
\newblock Rethinking semantic segmentation from a sequence-to-sequence
  perspective with transformers.
\newblock In {\em CVPR}, 2021.

\bibitem{CycleGAN2017}
Jun-Yan Zhu, Taesung Park, Phillip Isola, and Alexei~A Efros.
\newblock Unpaired image-to-image translation using cycle-consistent
  adversarial networks.
\newblock In {\em ICCV}, 2017.

\bibitem{zhu2021deformable}
Xizhou Zhu, Weijie Su, Lewei Lu, Bin Li, Xiaogang Wang, and Jifeng Dai.
\newblock Deformable detr: Deformable transformers for end-to-end object
  detection.
\newblock In {\em ICLR}, 2021.

\bibitem{zou2018unsupervised}
Yang Zou, Zhiding Yu, BVK Kumar, and Jinsong Wang.
\newblock Unsupervised domain adaptation for semantic segmentation via
  class-balanced self-training.
\newblock In {\em ECCV}, 2018.

\end{thebibliography}
}

\newpage
\clearpage
\renewcommand{\thesection}{S\arabic{section}}  
\renewcommand{\thefigure}{S\arabic{figure}}
\renewcommand{\thetable}{S\arabic{table}}
\setcounter{equation}{0}
\setcounter{figure}{0}
\setcounter{table}{0}
\setcounter{page}{1}
\setcounter{section}{0}

\section*{Supplementary}
%===============================================================================
In this supplementary material, we provide the additional information for,
\begin{itemize}
    \item[\textbf{S1}] discussion about limitations, 
    \item[\textbf{S2}] detailed datasets description involved in our experiments,
    \item[\textbf{S3}] detailed implementation and training details of our proposed framework,
    \item[\textbf{S4}] more experimental results and analysis.
\end{itemize}

\section{Limitations Analysis}
\noindent\textbf{Limitations.} Though our proposed method improves the OSUDA and OSDG performance by a large margin, it is still not saturated and yet to achieve the performance of fully supervised learning on the target domain. It is mainly constrained by the limited available knowledge on the target domain. One possible strategy to further improve the performance is to utilize other modal knowledge in the target domain, \eg the text information. 
More specifically, in OSDG problem, GTA$\rightarrow$ Foggy Cityscapes, the ``foggy" text can provide the guidance for target domain knowledge extraction and the adaptation to the target domain.

\section{Datasets Description}
\noindent\textbf{Cityscapes.} Cityscapes~\cite{cordts2016cityscapes} dataset is a real-world urban street scene dataset, which is collected from some European cities. Following~\cite{luo2020adversarial,wu2021style}, one randomly selected image from the training set is taken as the target domain for training in the one-shot unsupervised sim-to-real domain adaptation (OSUDA) setting, and 500 validation set images are used for testing. The original images are of resolution 2048$\times$1024, and are finely annotated with 19 semantic classes.

\noindent\textbf{GTA.} GTA~\cite{richter2016playing} dataset is a synthetic urban scene dataset, rendered from an open-world game engine. The scenes in GTAV are based on the Los Angeles city. It contains 24,966 images with the resolution of 1914$\times$1052, which are densely labeled with 19 semantic classes. The semantic annotations are compatible with Cityscapes. 

\noindent\textbf{SYNTHIA.} SYNTHIA~\cite{ros2016synthia} dataset is a synthetic dataset rendered from a virtual city. It consists of 9,400 photo-realistic images with resolution 1280$\times$760, coming with 16-class pixel-level semantic annotations, which is a subset of Cityscapes.

\noindent\textbf{Foggy Cityscapes.} Foggy Cityscapes~\cite{sakaridis2018semantic,sakaridis2018model} derives from the Cityscapes dataset, by simulating the fog on the real Cityscapes images. Thus, Foggy Cityscapes inherits the semantic annotations of the real, clear counterparts from Cityscapes. Under the one-shot sim-to-real domain generalization (OSDG) setting, 500 validation set foggy cityscapes images are used for testing, and one randomly selected web foggy image from Foggy Driving~\cite{sakaridis2018semantic} is used as the relevant domain image. 

\section{Framework Implementation and Training Details}
In order to construct the pseudo-target domain, 1) for OSUDA experiments, we stylize the images with MUNIT~\cite{huang2018multimodal}, setting the perceptual loss weight as 2.0, 2) for OSDG experiments, we adopt the frequency based Fourier Transform~\cite{yang2020fda}, with the $\beta$ as 0.05. For the semantic segmentation model training, in accordance with other DA methods~\cite{Tsai_adaptseg_2018,vu2019advent,hoyer2022daformer}, the Cityscapes, GTA images are resized to 1024$\times$512, 1280$\times$720, respectively. We adopt the AdamW~\cite{loshchilov2019decoupled} optimizer to train the model, with the learning rate as 6$\times$10$^{-5}$. Taking two random crops 512$\times$512 into each batch, the total training iteration is set as 80000. We implement the whole framework with PyTorch~\cite{NEURIPS2019_9015}.

\begin{table}[]
    \centering
    \resizebox{\linewidth}{!}{
    \begin{tabular}{c|cc}
    \toprule
    Setting & Ground Truth $\y_j^s$ & Pseudo-Label $\tilde{\y}_j^s$ \\
    \midrule
        GTA$\rightarrow$Cityscapes & 53.22 & 55.37\\
        SYNTHIA$\rightarrow$Cityscapes & 49.52 & 50.99\\
    \bottomrule
    \end{tabular}
    }
    \caption{\textbf{Ground Truth Label $\y_j^s$ \vs Pseudo-Label $\tilde{\y}_j^s$,} for intermediate domain randomization in Eq.~(\textcolor{red}{3}) of the main paper.}
    
    \label{tab:gtvspseudo}
\end{table}

\section{Additional Experimental Results}
In Sec.~\textcolor{red}{4} of the main paper, the quantitative and qualitative experimental results on GTA, SYNTHIA$\rightarrow$Cityscapes benchmarks for OSUDA, and GTA, SYNTHIA$\rightarrow$Foggy Cityscapes benchmarks for OSDG are shown. Here we provide additional experimental results to further demonstrate the effectiveness of our proposed approach.

\noindent\textbf{Comparison to Other SOTA Methods.} In Table~\textcolor{red}{1} and Table \textcolor{red}{4} of the main paper, the comparisons to other state-of-the-art (SOTA) methods for OSUDA and OSDG are provided. Corresponding to Table~\textcolor{red}{1} and Table~\textcolor{red}{4}, we show the detailed per class IoU performance in Table~\ref{tab:gta_city}, Table~\ref{tab:synthia_city}, Table~\ref{tab:gta_fog} and Table~\ref{tab:synthia_fog}. It is shown that our proposed approach significantly outperforms other SOTA methods, further verifying the effectiveness of our method for both OSUDA and OSDG.

\noindent\textbf{Ground Truth Label $\y_j^s$ \vs Pseudo-Label $\tilde{\y}_j^s$.} In Eq.~(\textcolor{red}{3}) of the main paper, we utilize the pseudo-label $\tilde{\y}_j^s$ instead of the ground truth label $\y_j^s$ corresponding to $\x_j^{s}$ for the intermediate domain randomization, to prevent overfitting to the source domain. In order to prove the validity of using the pseudo-label $\tilde{\y}_j^s$, we conduct the experiments under the OSUDA benchmarks, GTA$\rightarrow$ Cityscapes, SYNTHIA$\rightarrow$ Cityscapes, respectively. As shown in Table~\ref{tab:gtvspseudo}, the strategy using the pseudo-label $\tilde{\y}_j^s$ outperforms the one using the ground truth label $\y_j^s$, 55.37\%, 50.99\% \vs 53.22\%, 49.52\%, demonstrating the effectiveness of Eq.~(\textcolor{red}{3}) in the main paper.

\begin{table*}[]
    \centering%
    \resizebox{\linewidth}{!}{%
    \begin{tabular}{l|cccccccccccccccccccc|c}
    \toprule
    Method & Num$^{\#}$ & Road&SW&Build&Wall&Fence&Pole&TL&TS&Veg&Terrain&Sky&Person&Rider&Car&Truck&Bus&Train&MC&Bike& mIoU\\
    \midrule
    \midrule
    Source only & 0 & 75.8 & 16.8 & 77.2 & 12.5 & 21.0 & 25.5 & 30.1 & 20.1 & 81.3 & 24.6 & 70.3 & 53.8 & 26.4 & 49.9 & 17.2 & 25.9 & 6.5 & 25.3 & 36.0 & 36.6 \\
    \midrule
    AdaptSegNet~\cite{Tsai_adaptseg_2018}& 1 &77.7 & 19.2 & 75.5 & 11.7 & 6.4 & 16.8 & 18.2 & 15.4 & 77.1 & 34.0 & 68.5 & 55.3 & 30.9 & 74.5 & 23.7 & 28.3 & 2.9 & 14.4 & 18.9 & 35.2 \\
    CLAN~\cite{luo2019taking} & 1 & 77.1 & 22.7 & 78.6 & 17.0 & 14.8 & 20.5 & 23.8 & 12.0 & 80.2 & 39.5 & 74.3 & 56.6 & 25.2 & 78.1 & 29.3 & 31.2 & 0.0 & 19.4 & 16.7 & 37.7 \\
    ADVENT~\cite{vu2019advent}&1 & 76.1 & 15.1 & 76.6 & 14.4 & 10.8 & 17.5 & 19.8 & 12.0 & 79.2 & 39.5 & 71.3 & 55.7 & 25.2 & 76.7 & 28.3 & 30.5 & 0.0 & 23.6 & 14.4 & 36.1\\
    CBST~\cite{zou2018unsupervised}& 1& 76.1 & 22.2 & 73.5 & 13.8 & 18.8 & 19.1 & 20.7 & 18.6 & 79.5 & 41.3 & 74.8 & 57.4 & 19.9 & 78.7 & 21.3 & 28.5 & 0.0 & 28.0 & 13.2 & 37.1 \\
    CycleGAN~\cite{CycleGAN2017}&1 & 80.3 & 23.8 & 76.7 & 17.3 & 18.2 & 18.1 & 21.3 & 17.5 & 81.5 & 40.1 & 74.0 & 56.2 & 38.3 & 77.1 & 30.3 & 27.6 & 1.7 & 30.0 & 22.2 & 39.6 \\
    OST~\cite{benaim2018one}&1 & 84.3 & 27.6 & 80.9 & 24.1 & 23.4 & 26.7 & 23.2 & 19.4 & 80.2 & 42.0 & 80.7 & 59.2 & 20.3 & 84.1 & 35.1 & 39.6 & 1.0 & 29.1 & 23.2 & 42.3 \\
    FSDR~\cite{huang2021fsdr}& $\geq$15 & 89.3 & 40.5 & 79.1 & 26.3 & 27.8 & 29.3 & 33.7 & 29.0 & 83.0 & 27.7 & 76.0 & 57.8 & 27.5 & 81.0 & 32.3 & 42.4 & 16.8 & 21.0 & 30.2 & 44.8 \\
    DRPC\cite{yue2019domain}& $\geq$15 & - & - & - & - & - & - & - & - & - & - & - & - & - & - & - & - & - & - & - & 42.5\\
    SADG\cite{peng2022semantic}&0 &  - & - & - & - & - & - & - & - & - & - & - & - & - & - & - & - & - & - & - & 45.33\\
    RobNet~\cite{choi2021robustnet} & 0 &  - & - & - & - & - & - & - & - & - & - & - & - & - & - & - & - & - & - & - & 42.87\\
    WEDGE~\cite{kim2021wedge} & 1000 &  - & - & - & - & - & - & - & - & - & - & - & - & - & - & - & - & - & - & - & 43.60 \\
    IBN~\cite{pan2018IBN-Net} & 0 & - & - & - & - & - & - & - & - & - & - & - & - & - & - & - & - & - & - & - & 37.42 \\
    ASM~\cite{luo2020adversarial}& $\geq$10 & 86.2 & 35.2 & 81.4 & 24.2 & 25.5 & 31.5 & 31.5 & 21.9 & 82.9 & 30.5 & 80.1 & 57.3 & 22.9 & 85.3 & 43.7 & 44.9 & 0.0 & 26.5 & 34.9 & 44.5\\
    SMPPM\cite{wu2021style}& 1 & 85.0 & 23.2 & 80.4 & 21.3 & 24.5 & 30.0 & 32.0 & 26.7 & 83.2 & 34.8 & 74.0 & 57.3 & 29.0 & 77.7 & 27.3 & 36.5 & 5.0 & 28.2 & 39.4 & 42.8 \\
    \midrule
        Ours (R101)& 1 & 80.88 & 32.62 & 85.82 & 36.11 & 30.68 & 40.7 & 43.66 & 41.71 & 84.07 & 30.72 & 84.48 & 65.38 & 27.56 & 85.98 & 36.47 & 51.36 & 24.13 & 26.68 & 30.67 & \textbf{49.46} \\
        Ours (MiT-B5)& 1 & 83.41 & 35.30 & 87.11 & 44.79 & 32.27 & 42.53 & 50.19 & 52.47 & 87.99 & 46.09 & 90.43 & 66.71 & 25.55 & 88.64 & 50.32 & 50.77 & 44.54 & 34.36 & 38.58 & \textbf{55.37}\\
    \bottomrule
    \end{tabular}
    }
    \caption{\textbf{GTA$\rightarrow$Cityscapes: One-Shot Domain Adaptation.} Num$^{\#}$ represents the number of \textbf{real} images used for training, which are from the target domain or other auxiliary datasets, \eg, ImageNet~\cite{deng2009imagenet} and WikiArt~\cite{huang2017arbitrary}. The baseline methods adopt the ResNet-101 backbone.}
    \label{tab:gta_city}
\end{table*}

\begin{table*}[]
    \centering%
    \resizebox{\linewidth}{!}{%
    \begin{tabular}{l|ccccccccccccccccc|cc}
    \toprule
    Method& Num$^{\#}$ & Road&SW&Build&Wall$^*$&Fence$^*$&Pole$^*$&TL&TS&Veg&Sky&Person&Rider&Car&Bus&MC&Bike& mIoU$^*$ & mIoU\\
    \midrule
    \midrule
    Source only & 0 & 36.30 & 14.64 & 68.78 & 9.17 & 0.20 & 24.39 & 5.59 & 9.05 & 68.96 & 79.38 & 52.45 & 11.34 & 49.77 & 9.53 & 11.03 & 20.66 & 33.65 & 29.45\\
    \midrule
    AdaptSegNet~\cite{Tsai_adaptseg_2018}& 1 &64.1 & 25.6 & 75.3 & - & - & - & 4.7 & 2.7 & 77.0 & 70.0 & 52.2 & 20.6 & 51.3 & 22.4 & 19.9 & 22.3 & 39.1 & - \\
    CLAN~\cite{luo2019taking}& 1 & 68.3 & 26.9 & 72.2 & - & - & - & 5.1 & 5.3 & 75.9 & 71.4 & 54.8 & 18.4 & 65.3 & 19.2 & 22.1 & 20.7 & 40.4 & - \\
    ADVENT~\cite{vu2019advent}& 1 & 65.7 & 22.3 & 69.2 & - & - & - & 2.9 & 3.3 & 76.9 & 69.2 & 55.4 & 21.4 & 77.3 & 17.4 & 21.4 & 16.7 & 39.9 & - \\
    CBST~\cite{zou2018unsupervised}& 1 & 59.6 & 24.1 & 72.9 & - & - & - & 5.5 & 13.8 & 72.2 & 69.8 & 55.3 & 21.1 & 57.1 & 17.4 & 13.8 & 18.5 & 38.5 & - \\
    OST~\cite{benaim2018one}& 1 & 75.3 & 31.6 & 72.1 & - & - & - & 12.3 & 9.3 & 76.1 & 71.1 & 51.1 & 17.7 & 68.9 & 19.0 & 26.3 & 25.4 & 42.8 & - \\
    FSDR~\cite{huang2021fsdr}& $\geq$15& 69.3 & 34.9 & 77.6 & 7.9 & 0.2 & 29.4 & 16.3 & 19.2 & 72.3 & 76.3 & 56.7 & 22.1 & 80.6 & 41.5 & 19.1 & 29.3 & 47.3 & 40.8 \\
    DRPC~\cite{yue2019domain}& $\geq$15& - & - & - & - & - & - & - & - & - & - & - & - & - & - & - & - & - & 37.6\\
    SADG~\cite{peng2022semantic}& 0 & - & - & - & - & - & - & - & - & - & - & - & - & - & - & - & - & - & 40.87\\
    RobNet~\cite{choi2021robustnet} & 0 & - & - & - & - & - & - & - & - & - & - & - & - & - & - & - & - & - & 37.21\\
    WEDGE~\cite{kim2021wedge} & 1000 & - & - & - & - & - & - & - & - & - & - & - & - & - & - & - & - & - & 40.31\\
    IBN~\cite{pan2018IBN-Net} & 0 & - & - & - & - & - & - & - & - & - & - & - & - & - & - & - & - & - & 34.18 \\
    ASM~\cite{luo2020adversarial}& $\geq$10& 85.7 & 39.7 & 77.1 & 1.1 & 0.0 & 24.2 & 2.1 & 9.2 & 76.9 & 81.7 & 43.4 & 11.4 & 63.9 & 15.8 & 1.6 & 20.3 & 40.7 & 34.6\\
    SMPPM~\cite{wu2021style}& 1 & 79.3 & 35.3 & 75.9 & 5.6 & 16.6 & 29.8 & 25.4 & 22.7 & 79.9 & 76.8 & 54.6 & 23.5 & 60.2 & 23.9 & 21.2 & 36.6 & 47.3 & 41.4\\
    \midrule
    Ours(R101)& 1 & 82.53 & 33.83 & 77.75 & 12.61 & 0.78 & 34.18 & 30.8 & 34.42 &  79.75 & 82.43 & 55.42 & 30.71 & 72.5 & 28.44 & 15.89 & 47.76 & \textbf{51.72} & \textbf{44.99} \\
    Ours(MiT-B5)& 1 & 81.35 & 37.34 & 84.76 & 19.54 & 1.24 & 43.67 & 43.03 & 34.37 & 86.49 & 90.03 & 63.84 & 32.76 & 79.62 & 42.68 & 27.96 & 47.24 & \textbf{57.80} & \textbf{50.99}\\
    \bottomrule
    \end{tabular}
    }
    \caption{\textbf{SYNTHIA$\rightarrow$Cityscapes: One-Shot Domain Adaptation.} Num$^{\#}$ represents the number of \textbf{real} images used for training, which are from the target domain or other auxiliary datasets, \eg, ImageNet~\cite{deng2009imagenet} and WikiArt~\cite{huang2017arbitrary}. The baseline methods adopt the ResNet-101 backbone. mIoU$^*$ represents the 13 classes mIoU performance when removing the 3 classes denoted by $*$, as the common practice in~\cite{Tsai_adaptseg_2018,tranheden2021dacs,vu2019advent}.}
    \label{tab:synthia_city}
\end{table*}

\begin{table*}[t]
    \centering%
    \resizebox{\linewidth}{!}{%
    \begin{tabular}{l|cccccccccccccccccccc|c}
    \toprule
    Method& Num$^{\#}$ & Road&SW&Build&Wall&Fence&Pole&TL&TS&Veg&Terrain&Sky&Person&Rider&Car&Truck&Bus&Train&MC&Bike& mIoU\\
    \midrule
    \midrule
    Source only & 0 & 69.5 & 12.9 & 65.6 & 10.5 & 6.8 & 39.5 & 41.7 & 20.4 & 62.7 & 7.5 & 63.5 & 58.5 & 31.1 & 62.3 & 16.3 & 31.9 & 1.4 & 22.0 & 10.8 & 33.4 \\
    \midrule
    BDL~\cite{li2019bidirectional} & 0 & - & - & - & - & - & - & - & - & - & - & - & - & - & - & - & - & - & - & - & 30.3 \\
    FDA~\cite{yang2020fda} & 0 & - & - & - & - & - & - & - & - & - & - & - & - & - & - & - & - & - & - & - & 35.3 \\
    GASF\cite{Kundu_2021_ICCV} & 0 &  - & - & - & - & - & - & - & - & - & - & - & - & - & - & - & - & - & - & - & 38.3 \\ 
    BDL~\cite{li2019bidirectional} & 2975 & 89.6 & 37.6 & 65.4 & 19.5 & 14.4 & 23.2 & 22.7 & 25.5 & 48.9 & 35.7 & 39.7 & 50.7 & 29.8 & 79.1 & 27.9 & 32.8 & 0.2 & 18.5 & 30.4 & 36.3 \\
    CBST~\cite{zou2018unsupervised} & 2975 & 74.6 & 30.3 & 73.2 & 7.0 & 20.0 & 40.7 & 47.4 & 35.4 & 53.0 & 5.8 & 65.4 & 47.7 & 21.7 & 75.4 & 21.7 & 39.1 & 5.5 & 18.5 & 33.5 & 37.7 \\
    MLSL~\cite{iqbal2020mlsl} & 2975 & 81.5 & 33.6 & 76.6 & 7.9 & 23.1 & 41.1 & 47.5 & 35.9 & 52.0 & 6.1 & 64.9 & 54.1 & 27.8 & 81.2 & 16.5 & 37.7 & 1.5 & 17.4 & 36.7 & 39.1 \\
    FogAdapt~\cite{iqbal2022fogadapt} & 2975 & 85.1 & 31.7 & 76.7 & 16.5 & 20.3 & 41.2 & 46.2 & 34.9 & 70.8 & 9.1 & 63.8 & 53.9 & 26.2 & 81.5 & 22.0 & 38.0 & 5.9 & 19.0 & 36.3 & 41.0 \\
    ASM~\cite{luo2020adversarial} & 0 & 83.15 & 32.83 & 69.41 & 13.3 & 23.47 & 29.24 & 22.24 & 22.81 & 66.35 & 23.32 & 64.7 & 53.38 & 22.07 & 75.49 & 18.41 & 22.25 & 0.00 & 18.85 & 22.46 & 35.99 \\
    \midrule
        Ours (R101) & 0 & 79.58 & 21.68 & 78.68 & 18.18 & 22.17 & 30.53 & 35.31 & 17.33 & 77.50 & 22.82 & 67.80 & 61.87 & 32.39 & 80.57 & 24.21 & 25.90 & 0.00 & 23.17 & 22.15 & \textbf{39.05}\\
        Ours (MiT-B5) & 0 & 84.98 & 33.09 & 80.87 & 37.6 & 28.57 & 34.38 & 44.66 & 27.12 & 81.5 & 33.51 & 73.2 & 64.42 & 36.42 & 84.37 & 34.85 & 45.25 & 41.95 & 30.76 & 34.28 & \textbf{49.04}\\
    \bottomrule
    \end{tabular}
    }
    \caption{\textbf{GTA$\rightarrow$Foggy Cityscapes: One-Shot Domain Generalization.} Num$^{\#}$ represents the number of \textbf{Foggy Cityscapes} (target domain) images used for training. The baseline methods adopt the ResNet-101 backbone.}
    \label{tab:gta_fog}
\end{table*}

\begin{table*}[]
    \centering%
    \resizebox{\linewidth}{!}{%
    \begin{tabular}{l|ccccccccccccccccc|cc}
    \toprule
    Method& Num$^{\#}$ & Road&SW&Build&Wall$^*$&Fence$^*$&Pole$^*$&TL&TS&Veg&Sky&Person&Rider&Car&Bus&MC&Bike& mIoU$^*$ & mIoU\\
    \midrule
    \midrule
    Source only & 0 & 29.3 & 21.3 & 34.5 & 0.8 & 0.0 & 17.5 & 15.8 & 8.2 & 17.1 & 33.5 & 57.1 & 4.7 & 71.2 & 12.2 & 2.9 & 8.9 & 24.4 & 20.9 \\
    \midrule
    BDL~\cite{li2019bidirectional} & 2975 & 83.2 & 43.2 & 63.6 & 2.38 & 0.1 & 19.1 & 6.8 & 5.3 & 35.4 & 19.9 & 55.4 & 31.8 & 65.2 & 21.0 & 27.6 & 37.1 & 38.1 & 32.3 \\
    CBST~\cite{zou2018unsupervised} & 2975 & 70.5 & 31.2 &  57.62 & 2.9 & 0.02 & 31.7 & 29.1 & 23.1 & 38.5 & 41.1 & 61.3 & 18.9 & 75.0 & 8.3 & 11.9 & 32.1 & 38.4 & 33.3 \\
    MLSL~\cite{iqbal2020mlsl} & 2975 & 48.9 & 27.2 & 53.4 & 11.4 & 0.4 & 31.9 & 32.4 & 21.0 & 49.2 & 40.1 & 65.8 & 24.4 & 77.8 & 20.9 & 19.0 & 50.5 & 40.8 & 35.9\\
    FogAdapt~\cite{iqbal2022fogadapt} & 2975 & 62.2 & 28.0 & 56.4 & 13.1 & 0.7 & 30.3 & 30.1 & 27.4 & 61.7 & 61.8 & 54.9 & 30.0 & 66.1 & 2.6 & 12.1 & 44.8 & 41.4 & 36.4\\
    ASM~\cite{luo2020adversarial} & 0 & 75.39 & 29.04 & 67.85 & 1.79 & 0.05 & 21.54 & 4.88 & 8.51 & 59.78 & 58.46 & 37.83 & 11.91 & 49.47 & 7.58 & 3.17 & 16.68 & 33.12 & 28.37 \\
    \midrule
    Ours(R101) & 0 & 79.91 & 32.49 & 69.53 & 11.97 & 0.22 & 30.52 & 29.89 & 22.56 & 65.98 & 59.91 & 59.57 & 26.23 & 78.34 & 28.87 & 25.37 & 36.04 & \textbf{47.28} & \textbf{41.09}\\
    Ours(MiT-B5) & 0 & 74.5 & 30.2 & 78.21 & 16.48 & 1.89 & 37.93 & 35.17 & 24.87 & 77.47 & 69.03 & 61.14 & 25.45 & 81.96 & 37.97 & 24.69 & 31.56 & \textbf{50.17} & \textbf{44.28} \\
    \bottomrule
    \end{tabular}
    }
    \caption{\textbf{SYNTHIA$\rightarrow$Foggy Cityscapes: One-Shot Domain Generalization.} Num$^{\#}$ represents the number of \textbf{Foggy Cityscapes} (target domain) images used for training. The baseline methods adopt the ResNet-101 backbone. mIoU$^*$ represents the 13 classes mIoU performance when removing the 3 classes denoted by $*$, as the common practice in~\cite{Tsai_adaptseg_2018,tranheden2021dacs,vu2019advent}.}
    \label{tab:synthia_fog}
\end{table*}

\end{document}